    \documentclass{bmcart}

        \usepackage[utf8]{inputenc} 
        \usepackage{url}
    
        \usepackage{color}
        \usepackage{graphicx}
        \usepackage[justification=centering]{caption}
        \usepackage{CJK}
        \usepackage{amsmath}
        \usepackage[lined,boxed,commentsnumbered, ruled]{algorithm2e}
        \usepackage{algorithmic}
        \usepackage{epsfig}
        \usepackage{multirow}
        \usepackage{url}
        \usepackage{amssymb}
        
        \startlocaldefs
        \endlocaldefs

\begin{document}
        
        \begin{frontmatter}
        
        \begin{fmbox}
        \dochead{Research}
        
        
        \title{Clinical Text Classification with Rule-based Features and Knowledge-guided Convolutional Neural Networks}
        
        
        \author[
           addressref={aff1},                   
           email={liang.yao@northwestern.edu}   
        ]{\fnm{Liang} \snm{Yao}}
        \author[
           addressref={aff1},
           email={chengsheng.mao@northwestern.edu}
        ]{\fnm{Chengsheng} \snm{Mao}}
        \author[
           addressref={aff1},
           corref={aff1},   
           email={yuan.luo@northwestern.edu}
        ]{\fnm{Yuan} \snm{Luo}}        
        
        \address[id=aff1]{
          \orgname{Northwestern University}, 
          \city{Chicago, Illinois 60611},                              
          \cny{United States}                                    
        }
        \address[id=aff2]{%
        \orgname{Northwestern University}, 
        \city{Chicago, Illinois 60611},                              
        \cny{United States}       
        }
        
        
        \end{fmbox}
        
        
        \begin{abstractbox}
        
        \begin{abstract} 
            Clinical text classification is an important problem in medical natural language processing.
            Existing studies have conventionally focused on rules or knowledge sources-based feature engineering, but only a few have exploited effective feature learning capability of deep learning methods. 
            In this study, we propose a novel approach which combines rule-based features and knowledge-guided deep learning techniques for effective disease classification. 
            Critical Steps of our method include identifying trigger phrases, predicting classes with very few examples using trigger phrases and training a convolutional neural network with word embeddings and Unified Medical Language System (UMLS) entity embeddings.
            We evaluated our method on the 2008 Integrating Informatics with Biology and the Bedside (i2b2) obesity challenge.
            The results show that our method outperforms the state of the art methods.
        \end{abstract}
        
        
        
        
        \end{abstractbox}
        %
        
        \end{frontmatter}
        
        
        
        \section{Introduction}

        Clinical records are an important type of electronic health record (EHR) data and often contain valuable and detailed patient information and clinical experiences of doctors.
        As a fundamental task of natural language processing, text classification plays an important role in clinical records organization and retrieval, and it can support cohort identification and clinical decision~\cite{huang2015community,demner2009can}.
          
        Existing clinical text classification studies often use different forms of rules or knowledge sources for feature engineering~\cite{wilcox2003role,suominen2008machine,solt2009semantic,garla2013knowledge,garla2012ontology}. But most of the studies could not learn effective features automatically, while deep learning methods have shown powerful feature learning capability recently in the general domain~\cite{goodfellow2016deep}.
          
        In this work, we propose a new method which combines rule-based features and knowledge-guided deep learning techniques for disease classification. We first identify trigger phrases using rules, then use these trigger phrases to predict classes with very few examples, and finally train a convolutional neural network on the trigger phrases with word embeddings and Unified Medical Language System (UMLS)~\cite{bodenreider2004unified} Concept Unique Identifiers (CUIs) with entity embeddings. We evaluated our method on the 2008 Integrating Informatics with Biology and the Bedside (i2b2) obesity challenge~\cite{uzuner2008identifying}, a multilabel classification task focused on obesity and its 15 most common comorbidities (diseases). The results demonstrate that our method outperforms state of the art methods for the challenge.
          
          \section{Related Work}
          
        \subsection{Clinical Text Classification}
        A systematic literature review of clinical coding and classification systems has been conducted by Stanfill et al.~\cite{stanfill2010systematic}. 
        Some challenge tasks in biomedical text mining also focus on clinical text classification, 
        e.g., Informatics for Integrating Biology and the Bedside (i2b2) hosted text classification tasks on determining smoking status~\cite{uzuner2008identifying}, 
        and predicting obesity and its co-morbidities~\cite{uzuner2009recognizing}. In this work, we focus on the obesity challenge~\cite{uzuner2009recognizing}. Among the top ten systems of obesity challenge, most are rule-based systems, and the top four systems are purely rule-based.

        Many approaches for clinical text classification rely on biomedical knowledge sources~\cite{wilcox2003role}. 
        A common approach is to first map narrative text to concepts from knowledge sources like Unified Medical Language System (UMLS), 
        then train classifiers on document representations that include UMLS Concept Unique Identifiers (CUIs) as features~\cite{garla2013knowledge}. 
        More knowledge-intensive approaches enrich the feature set with related concepts~\cite{suominen2008machine} for apply semantic kernels that project documents that contain related concepts closer together in a feature space~\cite{garla2012ontology}. Similarly, Yao et al.~\cite{yao2016traditional} proposed to improve distributed document representations with medical concept descriptions for traditional Chinese medicine clinical records classification.
          
        On the other hand, some clinical text classification studies use various types of information instead of knowledge sources. 
        For instance, effective classifiers have been designed based on regular expression discovery~\cite{bui2014learning} and semi-supervised learning~\cite{wang2010semi,garla2013semi}. 
        Active learning~\cite{figueroa2012active} has been applied in clinical domain, which leverages unlabeled corpora to improve the classification of clinical text.
          
        Although these methods used rules, knowledge sources or different types of information in many ways. They seldom use effective feature learning methods, while deep learning methods are recently widely used for text classification and have shown powerful feature learning capabilities.

        \subsection{Deep Learning for Clinical Data Mining}
          
        Recently, deep learning methods have been successfully applied to clinical data mining. 
        Two representative deep models are convolutional neural networks (CNN)~\cite{kim2014convolutional,kalchbrenner2014convolutional} and recurrent neural networks (RNN)~\cite{tai2015improved,yang2016hierarchical}. They achieve state of the art performances on a number of clinical data mining tasks. Beaulieu-Jones et al.~\cite{beaulieu2016semi} developed a neural network approach to construct phenotypes to classify patient disease status. The model obtained better performance than SVM, random forest, and decision tree models. They also claimed to successfully learn the structure of high-dimensional EHR data for phenotype stratification. Gehrmann et al.~\cite{gehrmann2017comparing} compared convolutional neural networks to the traditional rule-based entity extraction systems using the cTAKES and logistic regression using n-gram features. They tested ten different phenotyping tasks using discharge summaries. The CNN outperformed other phenotyping algorithms in the prediction of ten phenotypes, and they concluded that NLP-based deep learning methods improved the performance of patient phenotyping compared to other methods. Luo et al. applied both CNN and RNN to classify the semantic relations between medical concepts in discharge summaries from the i2b2-VA challenge dataset~\cite{uzuner20112010} and showed that CNN and RNN with only word embedding features can obtain similar performances compared to state-of-the-art systems by challenge participants with heavy feature engineering~\cite{luo2017recurrent,luo2017segment}. Wu et al.~\cite{wu2015named} applied CNN using a set of pre-trained embeddings on clinical text for named entity recognization. They found that their models outperformed the baseline of conditional random fields (CRF). Geraci et al.~\cite{geraci2017applying} applied deep neural networks to identify youth depression from unstructured text notes. The authors achieved a sensitivity of 93.5\% and a specificity of 68\%. Jagannatha et al.~\cite{jagannatha2016structured,jagannatha2016bidirectional} experimented with RNN, long short-term memory (LSTM), gated recurrent units (GRU), bidirectional LSTM, combinations of LSTM with CRF, and CRF to extract clinical concepts from texts. They found that all variants of RNN outperformed the CRF base-line. Lipton et al.~\cite{lipton2015learning} evaluated the performance of LSTM in phenotype prediction using multivariate time series clinical measurements. They concluded that their model outperformed logistic regression and multi-layer perceptron (MLP). They also concluded that the combination of LSTM and MLP had the best performance. Che et al.~\cite{che2015deep} also applied deep learning methods to study time series in ICU data. They introduced a prior-based Laplacian regularization process on the sigmoid layer that is based on medical ontologies and other structured knowledge. In addition, they developed an incremental training procedure to iteratively add neurons to the hidden layer. Then they applied causal inference techniques to analyze and interpret the hidden layer representations. They demonstrated that their proposed methods improved the performance of phenotype identification and that the model trains with faster convergence and better interpretation. 
        
        Although deep learning techniques have been well studied in clinical data mining, most of these works do not focus on long clinical text classification (e.g., an entire clinical note) or utilize knowledge sources, while we propose a novel knowledge-guided deep learning method for clinical text classification. 
        
        \section{Obesity Challenge}
          
        The objective of the i2b2 2008 obesity challenge~\cite{uzuner2009recognizing} is to assess text classification methods for determining patient disease status with respect to obesity and 15 of its comorbidities: Diabetes mellitus (DM), Hypercholesterolemia, Hypertriglyceridemia, Hypertension, atherosclerotic cardiovascular disease (CAD), Heart failure (CHF), Peripheral vascular disease (PVD), Venous insufficiency, Osteoarthritis (OA), Obstructive sleep apnea (OSA), Asthma, Gastroesophageal reflux disease (GERD), Gallstones, Depression, and Gout. Our goal is to label each document as either Present (Y), Absent (N), Questionable (Q) or Unmentioned (U) for each disease. Macro $F_1$ score is the primary metric for evaluating and ranking classification methods.
          
        The challenge consists of two tasks, namely textual task and intuitive task. The textual task is to identify explicit evidences of the diseases, while the
        intuitive task focused on the prediction of the disease status when the evidence is not explicitly mentioned. Thus, the Unmentioned (U) class label was excluded from the intuitive task. The classes are distributed very unevenly: there are only few N and Q examples in textual task data set and few Q examples in intuitive task data set, as shown in Table 1. There exist classes even without training example. For instance,  there is no training  example with Q and N label for Depression in textual task, and there is no training  example with Q label for Gallstones in intuitive task. The details of the datasets can be found in~\cite{uzuner2009recognizing}. 
          
          {\small
          \begin{table}[h]\footnotesize
          \centering
          \renewcommand{\arraystretch}{1.2}
          \label{distribution}
          \caption{The class distribution in the obesity challenge datasets.}
          \begin{tabular}{c|c|c|c|c}
          \hline
          \multirow{2}*{Label}&	\multicolumn{2}{c|}{Training Set} & \multicolumn{2}{c}{Test Set}\\
          \cline{2-5}
           & Textual & Intuitive & Textual & Intuitive\\
          \hline
          Y & 3208 & 3267 & 2192& 2285\\
          \hline
           N & 87 & 7362 & 65 & 5100\\
          \hline
           Q & 39 & 26 & 17 & 14\\
          \hline
          U  & 8296 & 0 & 5770 & 0\\
          \hline
          \end{tabular}
          \end{table}
          }
          
        \section{Method}
          
        Our method consists of three steps: identifying trigger phrases, predicting classes with very few examples using trigger phrases and training a knowledge-guided convolutional neural network for more populated classes\footnote{We released the implementation at \url{https://github.com/yao8839836/obesity}.}. We base our method on Solt's system~\cite{solt2009semantic} to identify trigger phrases and predict classes with very few examples. Solt's system is a very strong rule-based system. It ranked the first place in the intuitive task and the second place in the textual task and overall first place in i2b2 Obesity challenge. Solt's system can discover very informative trigger phrases with positive, negative or uncertain contexts. We use a Perl implementation\footnote{\url{https://github.com/yao8839836/obesity/tree/master/perl_classifier}} of Solt's system provided by the authors.
          
        \subsection{Trigger Phrases Identification}
          
        We follow Solt's system~\cite{solt2009semantic} to identify trigger phrases. We first do the same preprocessing including abbreviation resolution and family history removing. We then use the same disease names, their directly associated terms and negative/uncertain words to identify trigger phrases. The trigger phrases are disease names (e.g., Gallstones) and their alternative names (e.g., Cholelithiasis) with or without negative/uncertain words.

        \subsection{Predicting Classes with Very Few Examples using Trigger Phrases}
          
        As mentioned above, the classes in obesity challenge are very unbalanced, and some classes even don't have training examples. Therefore we could not predict these classes using machine learning methods and resort to rules defined in Solt's system~\cite{solt2009semantic}. We exclude classes with very few examples in training set of each disease. Specifically, we remove examples with Q or N label for textual task and remove examples with Q label in intuitive task. Then for examples in the test set, we use trigger phrases to predict their labels. Following Solt's system~\cite{solt2009semantic}, we assume positive trigger phrases (disease names/alternatives without negative/uncertain words) are prior to negative trigger phrases, and negative trigger phrases are prior to uncertain trigger phrases. Therefore, if a clinical record has uncertain trigger phrases and dose not have positive/negative trigger phrases, we label it as Q. Similarly, if a clinical record has negative trigger phrases and dose not have positive trigger phrases, we label it as N.
          
        \subsection{Knowledge-guided Convolutional Neural Networks}
          
        \begin{figure*}[t]
        \centering
        \includegraphics[height=95 mm]{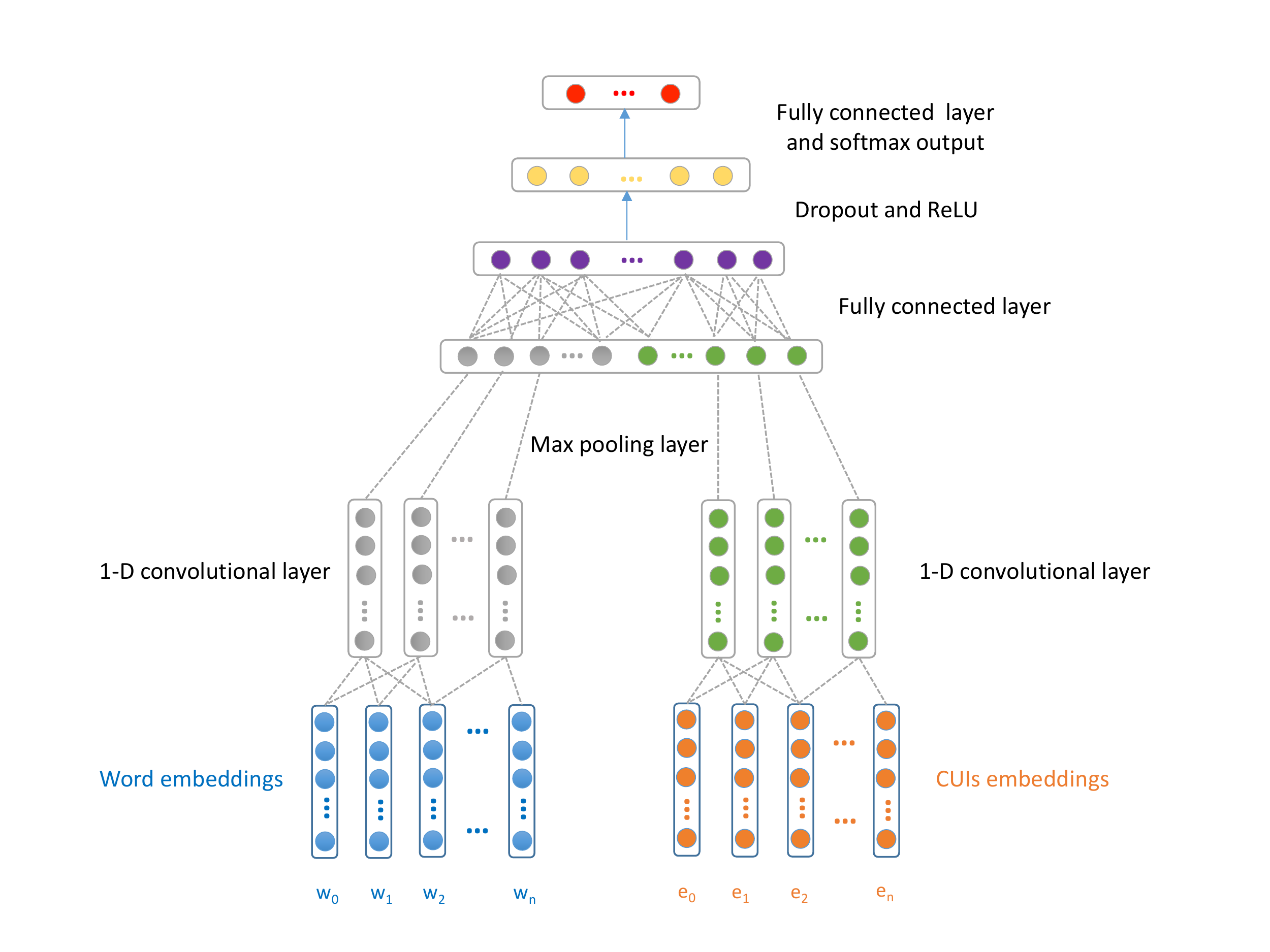}
        \caption{Our knowledge-guided convolutional neural network architecture.}
        \end{figure*}

    After removing classes with very few examples, there are only two classes in the training set of each disease (Y and U for textual task, Y and N for intuitive task). We train a Convolutional Neural Network (CNN) on positive trigger phrases and UMLS CUIs of training records, and classify test examples using the learned CNN model. CNN is a powerful deep learning model for text classification, and it performs better than recurrent neural networks in our preliminary experiment. The test phase of our method is given in Figure 2. If a record in test set is labeled Q or N by Solt's system, we trust Solt's system. Otherwise, we use the CNN to predict the label of the record.

        \begin{figure}[h]
        \centering
        \includegraphics[height=45 mm]{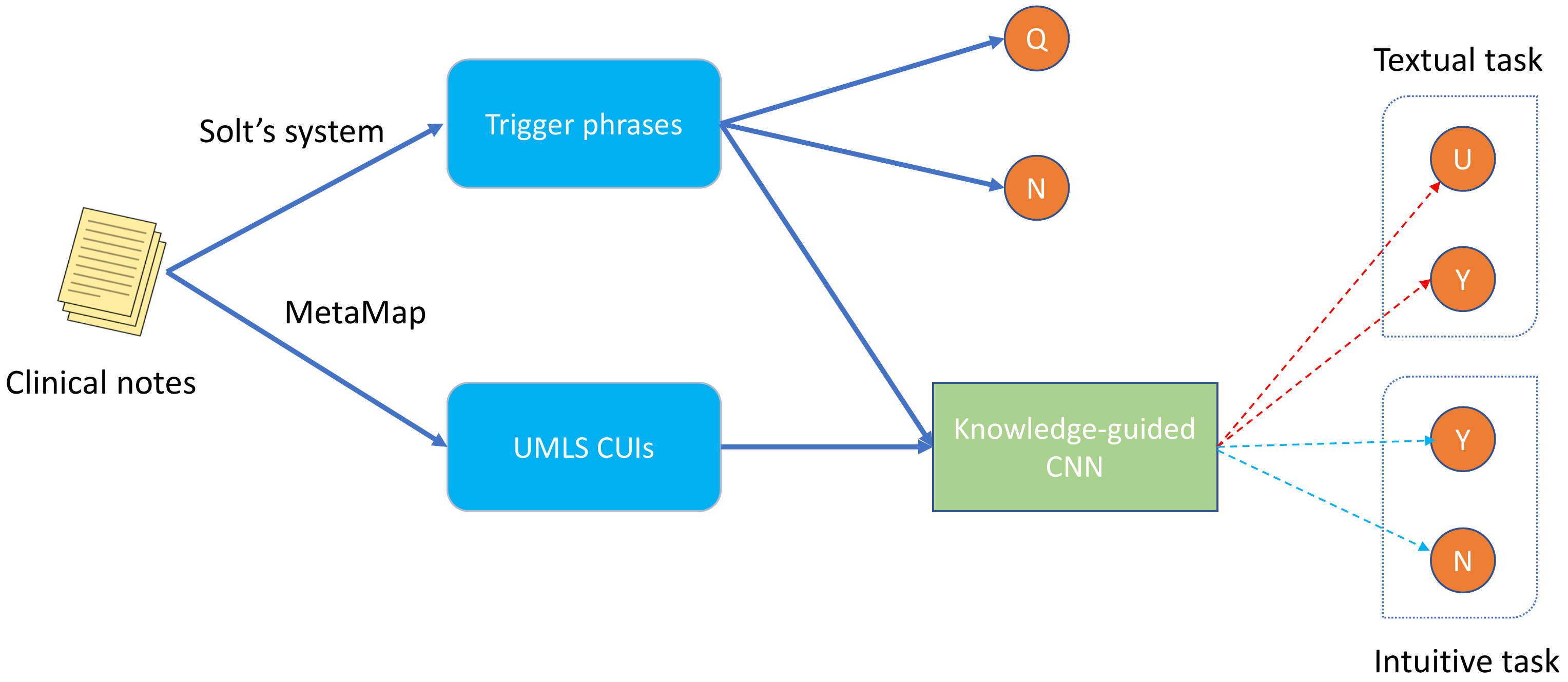}
        \caption{The test phase of our method.}
        \end{figure}
          
        For each disease, we use its positive trigger phrases with word2vec~\cite{mikolov2013distributed} word embeddings as the input of CNN. We used the pre-trained 200 dimensional word embeddings learned from MIMIC-III \cite{johnson2016mimic} clinical notes. We tried word embeddings with 100, 200, 300, 400, 500 and 600 dimensions, and found using 200 dimensional word embeddings performs the best. 
          
        We also use medical knowledge source to enrich the input of CNN model. We use MetaMap~\cite{aronson2010overview} to link the full clinical text to CUIs in UMLS~\cite{bodenreider2004unified}. After entity linking, a clinical record is represented as a bag of CUIs. We choose the following types of CUIs which are closely related to diseases as the input entities of CNN: Body Part, Organ, or Organ Component (T023), Finding (T033), Laboratory or Test Result (T034), Disease or Syndrome (T047), Mental or Behavioral Dysfunction (T048), Cell or Molecular Dysfunction (T049), Laboratory Procedure (T059), Diagnostic Procedure (T060), Therapeutic or Preventive Procedure (T061), Pharmacologic Substance (T121), Biomedical or Dental Material (T122), Biologically Active Substance (T123) and Sign or Symptom (T184). We list these CUIs types with type unique identifier (TUI) in Table 2. We found using the subset of CUIs leads to better results than using all CUIs. We use pre-trained CUIs embeddings made by De Vine et al~\cite{de2014medical} as the input entity representations of CNN.

        \begin{table}[h]\footnotesize
        \centering
        \renewcommand{\arraystretch}{1.2}
        \label{table_1}
        \caption{The types of CUIs we used.}
        \begin{tabular}{c|c}
        \hline
         TUI & Semantic type description\\
        \hline
        T023& Body Part, Organ, or Organ Component\\
        \hline
        T033  & Finding\\
        \hline
        T034  &  Laboratory or Test Result\\
        \hline
        T047  & Disease or Syndrome\\
        \hline
        T048 &  Mental or Behavioral Dysfunction\\
        \hline
        T049 &  Cell or Molecular Dysfunctions\\
        \hline
        T059 &   Laboratory Procedure\\
        \hline
        T060 & Diagnostic Procedure\\
        \hline
        T061 & Therapeutic or Preventive Procedure\\
        \hline
        T121 & Pharmacologic Substance\\
        \hline
        T122 & Biomedical or Dental Material\\
        \hline
        T123 & Biologically Active Substance\\
        \hline
        T184 & Sign or Symptom\\
        \hline
        \end{tabular}
        \end{table}

        Our CNN architecture is given in Fig. 1. The input layer consists of word embeddings of positive trigger phrases and CUIs embeddings of selected CUIs in each clinical record. $w_0, w_1, w_2, \ldots, w_n$
        are words in positive trigger phrases and $e_0, e_1, e_2, \ldots, e_n$ are CUIs in a record. A one-dimensional convolution layer is built on the input word embeddings and entity embeddings. We use max pooling to select the most important feature with the highest value in the convolutional feature map. We then concatenate the max pooling results of word embeddings and CUIs embeddings. The concatenated hidden features are fed into a fully connected layer, then a dropout and ReLU activation layer. Finally, a fully connected layer is fed to a softmax output layer, whose output is the probability distribution over labels.
          
        We implement our CNN model using TensorFlow~\cite{abadi2016tensorflow}, a popular deep learning framework. We set the following parameters for our model: the number of convolution filters: 256, the convolution kernel size: 5, the dimension of hidden layer in the fully connected layer: 128, dropout keep probability: 0.8, learning rate: 0.001, batch size: 64, the number of learning epochs: 30. We also tried other settings of these parameters but do not find much difference. We use softmax cross entropy loss as the loss function and Adam algorithm~\cite{kinga2015method} as the optimizer.

          \begin{table*}[h]\scriptsize
          \centering
          \renewcommand{\arraystretch}{1.1}
          \label{table_1}
          \caption{Macro $F_1$ scores and Micro $F_1$ scores of Solt's system~\cite{solt2009semantic} (paper) and our method with word and entity embeddings. Scores in bold font means they are higher than the corresponding scores of the paper and Perl implementation.}
          \begin{tabular}{c|c|c|c|c|c|c|c|c}
          \hline
          \multirow{3}*{Disease} &	\multicolumn{4}{c|}{Solt's paper~\cite{solt2009semantic}}& \multicolumn{4}{c}{Our method with word \& entity embeddings}\\
          \cline{2-9}
          &	\multicolumn{2}{|c}{Textual}& \multicolumn{2}{|c}{Intuitive}&	\multicolumn{2}{|c}{Textual}& \multicolumn{2}{|c}{Intuitive}\\
          \cline{2-9}
          & Macro $F_1$ &Micro $F_1$&Macro $F_1$&Micro $F_1$&Macro $F_1$ &Micro $F_1$&Macro $F_1$&Micro $F_1$  \\
          \hline 
          Asthma& 0.9434 &0.9921 &0.9784& 0.9894&0.9434 & 0.9921 & 0.9784& 0.9894\\
          CAD & 0.8561& 0.9256& 0.6122& 0.9192&0.8551 & 0.9235 & \textbf{0.6233} & \textbf{0.9345}\\
          CHF & 0.7939& 0.9355& 0.6236& 0.9315&0.7939& 0.9355 & 0.6236& 0.9315\\
          Depression  & 0.9716& 0.9842& 0.9346& 0.9539& 0.9716& 0.9842 & \textbf{0.9602} & \textbf{0.9727}\\
          DM  & 0.9032& 0.9761& 0.9682& 0.9729& 0.9056 &0.9801 & 0.9731& 0.9770\\
          Gallstones  & 0.8141& 0.9822& 0.9729& 0.9857& 0.8141& 0.9822 &0.9689 & 0.9837\\
          GERD  & 0.4880& 0.9881& 0.5768& 0.9131& 0.4880& 0.9881 & 0.5768& 0.9131\\
          Gout  & 0.9733& 0.9881& 0.9771& 0.9900& 0.9733& 0.9881 & 0.9771& 0.9900\\
          Hypercholesterolemia  & 0.7922& 0.9721& 0.9053& 0.9072& 0.7922& 0.9721 &\textbf{0.9113} &0.9118\\
          Hypertension  & 0.8378& 0.9621& 0.8851& 0.9283& 0.8378& 0.9621& \textbf{0.9240} & \textbf{0.9484}\\
          Hypertriglyceridemia  & 0.9732& 0.9980& 0.7981& 0.9712& 0.9434 & 0.9961& 0.7092 & 0.9630\\
          OA  & 0.9594& 0.9761& 0.6286& 0.9589& 0.9626& 0.9781& 0.6307 & 0.9610\\
          Obesity  & 0.4879& 0.9675& 0.9724& 0.9732& 0.4885 &0.9696 &0.9747 & 0.9754\\
          OSA  & 0.8781& 0.9920& 0.8805& 0.9939& 0.8781& 0.9920 & 0.8805& 0.9939\\
          PVD & 0.9682& 0.9862& 0.6348& 0.9763& 0.9682& 0.9862 & 0.6314 & 0.9742\\
          Venous insufficiency & 0.8403& 0.9822& 0.8083 &0.9625&\textbf{0.8816}&\textbf{0.9882}&0.8083 &0.9625\\
          \hline
          Overall & 0.8000& 0.9756& 0.6745& 0.9590&\textbf{0.8016} & \textbf{0.9763} &\textbf{0.6768} &\textbf{0.9624}\\
          \hline
          \end{tabular}
          \end{table*}

        \begin{table*}[h]\scriptsize
        \centering
        \renewcommand{\arraystretch}{1.1}
        \label{table_1}
        \caption{Macro $F_1$ scores and Micro $F_1$ scores of Solt's system~\cite{solt2009semantic} (code) and our method with word embeddings only. Scores in bold font means they are higher than the corresponding scores of the paper and Perl implementation.}
        \begin{tabular}{c|c|c|c|c|c|c|c|c}
        \hline
        \multirow{3}*{Disease}&	\multicolumn{4}{c|}{Solt's code}& \multicolumn{4}{c}{Our method with word embeddings only}\\
        \cline{2-9}
        &	\multicolumn{2}{|c}{Textual}& \multicolumn{2}{|c}{Intuitive}&	\multicolumn{2}{|c}{Textual}& \multicolumn{2}{|c}{Intuitive}\\
        \cline{2-9}
        & Macro $F_1$ &Micro $F_1$&Macro $F_1$&Micro $F_1$&Macro $F_1$ &Micro $F_1$&Macro $F_1$&Micro $F_1$  \\
        \hline 
            Asthma& 0.9434 &0.9921 &0.9784& 0.9894&0.9434 & 0.9921 & 0.9784& 0.9894\\
            CAD & 0.8551& 0.9235& 0.6122& 0.9192& 0.8551& 0.9235 & 0.6122& 0.9192\\
            CHF & 0.7939& 0.9355& 0.6236& 0.9315& 0.7939& 0.9355 & 0.6236& 0.9315\\
            Depression  & 0.9716& 0.9842& 0.9346& 0.9539& 0.9716& 0.9842 &\textbf{0.9602}& \textbf{0.9767}\\
            DM  & 0.9056& 0.9801& 0.9731& 0.9770& 0.9056& 0.9801 & 0.9731& 0.9770\\
            Gallstones  & 0.8141& 0.9822& 0.9729& 0.9857& 0.8141& 0.9822 &0.9729 &0.9857\\
            GERD  & 0.4880& 0.9881& 0.5768& 0.9131& 0.4880& 0.9881 & 0.5768& 0.9131\\
            Gout  & 0.9733& 0.9881& 0.9771& 0.9900& 0.9733& 0.9881 & 0.9771& 0.9900\\
            Hypercholesterolemia  & 0.7922& 0.9721& 0.9101& 0.9118& 0.7922& 0.9721 &0.9042 & 0.9049\\
            Hypertension  & 0.8378& 0.9621& 0.8861& 0.9283& 0.8378& 0.9621 & \textbf{0.9240}& \textbf{0.9484}\\
            Hypertriglyceridemia  & 0.9732& 0.9980& 0.7092& 0.9630& 0.9732& 0.9980 & 0.7092& 0.9630\\
            OA  & 0.9626 & 0.9781& 0.6307& 0.9610& 0.9626 & 0.9781 & 0.6307& 0.9610\\
            Obesity  & 0.4885& 0.9696 & 0.9747& 0.9754& 0.4885& 0.9696 & 0.9747& 0.9754\\
            OSA  & 0.8781& 0.9920& 0.8805& 0.9939& 0.8781& 0.9920 & 0.8805& 0.9939\\
            PVD & 0.9682& 0.9862& 0.6314& 0.9742& 0.9682& 0.9862 & 0.6314& 0.9742\\
            Venous insufficiency & 0.8403& 0.9822& 0.8083 &0.9625& 0.8403& 0.9822 &0.8083 &0.9625\\
        \hline
            Overall & 0.8014& 0.9760& 0.6745& 0.9592& 0.8014 & 0.9760 & \textbf{0.6760}   & \textbf{0.9612}\\
        \hline
        \end{tabular}
        \end{table*}

        \begin{table*}[h]\scriptsize
            \centering
            \renewcommand{\arraystretch}{1.1}
            \label{table_1}
            \caption{Macro $F_1$ scores and Micro $F_1$ scores of Logistic Regression and SVM. Classes with very few examples are labeled by Solt's system.}
            \begin{tabular}{c|c|c|c|c|c|c|c|c}
            \hline
            \multirow{3}*{Disease}&	\multicolumn{4}{c|}{Logistic Regression}& \multicolumn{4}{c}{SVM}\\
            \cline{2-9}
            &	\multicolumn{2}{|c}{Textual}& \multicolumn{2}{|c}{Intuitive}&	\multicolumn{2}{|c}{Textual}& \multicolumn{2}{|c}{Intuitive}\\
            \cline{2-9}
            & Macro $F_1$ &Micro $F_1$&Macro $F_1$&Micro $F_1$&Macro $F_1$ &Micro $F_1$&Macro $F_1$&Micro $F_1$  \\
            \hline 
                Asthma & 0.9434 & 0.9921 & 0.9784 & 0.9894 & 0.9434& 0.9921 & 0.9784&0.9894 \\
                CAD & 0.8551 & 0.9235 & 0.6204 & 0.9301 & 0.8551& 0.9235 &0.6122 &0.9192\\
                CHF & 0.7939 & 0.9355 & 0.6236 & 0.9315 &0.7939 & 0.9355 & 0.6236& 0.9315\\
                Depression  & 0.9716 & 0.9842 & 0.9573& 0.9706 & 0.9716 & 0.9842 & 0.9573 & 0.9706\\
                DM  & 0.9056 & 0.9801 & 0.9731 & 0.9770 &0.9056 & 0.9801 & 0.9731& 0.9770\\
                Gallstones  & 0.8141  & 0.9822 & 0.9729 & 0.9857 &0.8141 & 0.9822 &0.9729 & 0.9857\\
                GERD  & 0.4880 & 0.9881 & 0.5768 & 0.9131&0.4880 & 0.9881 & 0.5768 & 0.9131\\
                Gout  & 0.9733 & 0.9881 & 0.9771& 0.9900 & 0.9733& 0.9881  &0.9771 & 0.99\\
                Hypercholesterolemia  &  0.7922 & 0.9721 & 0.9043 & 0.9049& 0.7922& 0.9721 &0.9134 &0.9142\\
                Hypertension  & 0.8378 & 0.9621 & 0.9271& 0.9507&0.8378 & 0.9621 &0.9271 & 0.9507\\
                Hypertriglyceridemia  & 0.9732 & 0.9980 &0.7092 & 0.9630 & 0.9732 & 0.9980 &0.7092 & 0.9630\\
                OA  &  0.9626& 0.9781 & 0.6307 & 0.961 &0.9626 & 0.9781 & 0.6307& 0.9610\\
                Obesity  & 0.4885 & 0.9696 & 0.9747 & 0.9754 &0.4885 &  0.9696 &0.9747 & 0.9754\\
                OSA & 0.8781  & 0.992 & 0.8805 & 0.9939 & 0.8781 & 0.9920 & 0.8805& 0.9939\\
                PVD & 0.9682 & 0.9862 & 0.6314 & 0.9742 & 0.9682 & 0.9862 & 0.6314& 0.9742\\
                Venous insufficiency & 0.8403 & 0.9822 & 0.8083 & 0.9625 & 0.8403 &0.9822  & 0.8083& 0.9625\\
            \hline
            \label{table_1}
                Overall & 0.8014& 0.9760 & 0.6764 & 0.9619 & 0.8014 & 0.9760 &  0.6764  & 0.9618\\
            \hline
            \end{tabular}
            \end{table*}
              
            \section{Results}
          
        Table 3 and Table 4 show Macro $F_1$ scores and Micro $F_1$ scores of our method and Solt's system. We report results of both the Solt's paper~\cite{solt2009semantic} and the Perl implementation because we base our method on the Perl implementation and we found there are some differences between the paper's results and Perl implementation's results. This is likely due to further feature engineering that are not reflected when Solt et al. submitted classification output to the challenge. For completeness of the results, we show the performances from both Solt's paper and code. We also report the results of our method when using only word embeddings as CNN input.

        From the two tables, we can note that the Perl implementation performs slightly better than the paper, the authors might not submit their best results to the obesity challenge. We can also observe that CNN with word embeddings performs better than the Perl implementation in intuitive task, which means using a deep learning model can learn effective features for better classification. The input trigger phrases for CNN are the same as the trigger phrases for Y/U (textual task) or Y/N (intuitive task) labeling in the Perl code.  The results in the textual task are not improved when using word embeddings only, because the textual task needs explicit evidences to label the records, and the positive trigger phrases contain enough information, therefore CNN with word embeddings may not be particularly helpful. Nevertheless, after adding CUIs embeddings as additional input, more scores for different diseases are improved, and the overall $F_1$ scores are higher than Solt's system in both textual task and intuitive task. This is likely due to the fact that the disambiguated CUIs are closely related to diseases and their embeddings contain more semantic information, which is helpful for disease classification. To the best of our knowledge, we have achieved the best results in intuitive task so far.

        Note that the results of Solt's paper and Perl implementation remain the same, while our method produces slightly different results in different runs. We run our model ten times and found the overall Macro $F_1$ scores and Micro $F_1$ scores are significantly higher ($p$ value $<$ 0.05 based on student $t$ test) than Solt's paper and implementation. We checked the cases our method failed to predict correctly. and found the most error cases are caused by using Solt's positive trigger phrases. For many error cases, our method predicted N or U when no positive trigger phrases are identified, but the real labels are Y. For some other cases, our method predicted Y when positive trigger phrases are identified, but the real labels are N or U.
        For some diseases, our proposed method and Solt's system achieved a very high Micro $F_1$ but a low Macro $F_1$. This is due to the fact that there are only a few Q or N records for these diseases (i.e., imbalanced class ratio), and we could not identify effective negative/uncertain trigger phrases using Solt's rules. The regular expressions in Solt's system can be further enriched so that we can identify trigger phrases more accurately.
        We note that the knowledge features part does not improve much. In fact, we think MetaMap will indeed introduce some noisy and unrelated CUIs, as previous studies also showed. To remedy this, following Weng et al.~\cite{weng2017medical}, we only kept CUIs from selected semantic types that are considered most relevant to clinical tasks. We found that filtering CUIs based on semantic types did lead to moderate performance improvement over using all CUIs. In another related computational phenotyping study~\cite{zeng2017contralateral}, we found that manually curated  CUI set resulted in significant performance improvement. We believe that improving entity recognition and integrating word/entity sense disambiguation will improve the performance, and plan to explore such directions in future work.

        We also compared our method with two commonly used classifiers: Logistic Regression and linear kernel support Vector Machine (SVM). We use LogisticRegression and LinearSVC class in scikit-learn as our implementations. For fair comparison, we use the same training set as knowledge-guided CNN. We represent a record as a binary vector, each dimension means whether an unique word is in its positive trigger phrases.  For test examples, we also use Solt's system to predict Q and N. If a test example is not labeled Q or N by Solt's system, we use Logistic Regression or SVM to predict the label. Table 5 shows the results, we can observe that the results are similar to our method with word embeddings only, which means positive trigger phrases themselves are informative enough, while word embeddings could not help to improve the performances. Nevertheless, we run our model ten times and found the overall Macro $F_1$ scores and Micro $F_1$ scores are significantly higher ($p$ value $<$ 0.05 based on student $t$ test) than SVM and Logistic Regression, which 
        verifies the effectiveness of CUIs embeddings again.
          
        \section{Conclusion}
        
        In this study, we present a new clinical text classification method which combines rule-based features and knowledge-guided deep learning techniques. Specifically, we use rules to identify trigger phrases which contain diseases names, their alternative names and negative/uncertain words, then use these trigger phrases to predict classes with very few examples, and finally train a knowledge-guided convolutional neural network (CNN) with word embeddings and UMLS CUIs embeddings. The evaluation results on the i2b2 obesity challenge show that our method outperforms the state of the art methods for the challenge. We showed that CNN model is powerful for learning effective hidden features, and CUIs embeddings are helpful for building clinical text representations. This shows integrating domain knowledge into CNN models is promising. We plan to develop more principled methods and evaluate the methods on more clinical records datasets in our future work.

\begin{backmatter}
\section*{Competing interests}
The authors declare that they have no competing interests.
            
\section*{Author's contributions}
Text for this section \ldots

\section*{Acknowledgment}

We would like to thank i2b2 National Center for Biomedical Computing funded by U54LM008748, for providing the clinical records originally prepared for the Shared Tasks for Challenges in NLP for Clinical Data organized by Dr. Ozlem Uzuner. We thank Dr. Uzuner for helpful discussions. We would like to also thank NVIDIA GPU Grant program for providing the GPU used in our computation. This work was supported in part by NIH Grant 1R21LM012618-01.

        
    \bibliographystyle{bmc-mathphys} 
    \bibliography{jbi}      


\begin{thebibliography}{40}
\ifx \bisbn   \undefined \def \bisbn  #1{ISBN #1}\fi
\ifx \binits  \undefined \def \binits#1{#1}\fi
\ifx \bauthor  \undefined \def \bauthor#1{#1}\fi
\ifx \batitle  \undefined \def \batitle#1{#1}\fi
\ifx \bjtitle  \undefined \def \bjtitle#1{#1}\fi
\ifx \bvolume  \undefined \def \bvolume#1{\textbf{#1}}\fi
\ifx \byear  \undefined \def \byear#1{#1}\fi
\ifx \bissue  \undefined \def \bissue#1{#1}\fi
\ifx \bfpage  \undefined \def \bfpage#1{#1}\fi
\ifx \blpage  \undefined \def \blpage #1{#1}\fi
\ifx \burl  \undefined \def \burl#1{\textsf{#1}}\fi
\ifx \doiurl  \undefined \def \doiurl#1{\textsf{#1}}\fi
\ifx \betal  \undefined \def \betal{\textit{et al.}}\fi
\ifx \binstitute  \undefined \def \binstitute#1{#1}\fi
\ifx \binstitutionaled  \undefined \def \binstitutionaled#1{#1}\fi
\ifx \bctitle  \undefined \def \bctitle#1{#1}\fi
\ifx \beditor  \undefined \def \beditor#1{#1}\fi
\ifx \bpublisher  \undefined \def \bpublisher#1{#1}\fi
\ifx \bbtitle  \undefined \def \bbtitle#1{#1}\fi
\ifx \bedition  \undefined \def \bedition#1{#1}\fi
\ifx \bseriesno  \undefined \def \bseriesno#1{#1}\fi
\ifx \blocation  \undefined \def \blocation#1{#1}\fi
\ifx \bsertitle  \undefined \def \bsertitle#1{#1}\fi
\ifx \bsnm \undefined \def \bsnm#1{#1}\fi
\ifx \bsuffix \undefined \def \bsuffix#1{#1}\fi
\ifx \bparticle \undefined \def \bparticle#1{#1}\fi
\ifx \barticle \undefined \def \barticle#1{#1}\fi
\ifx \bconfdate \undefined \def \bconfdate #1{#1}\fi
\ifx \botherref \undefined \def \botherref #1{#1}\fi
\ifx \url \undefined \def \url#1{\textsf{#1}}\fi
\ifx \bchapter \undefined \def \bchapter#1{#1}\fi
\ifx \bbook \undefined \def \bbook#1{#1}\fi
\ifx \bcomment \undefined \def \bcomment#1{#1}\fi
\ifx \oauthor \undefined \def \oauthor#1{#1}\fi
\ifx \citeauthoryear \undefined \def \citeauthoryear#1{#1}\fi
\ifx \endbibitem  \undefined \def \endbibitem {}\fi
\ifx \bconflocation  \undefined \def \bconflocation#1{#1}\fi
\ifx \arxivurl  \undefined \def \arxivurl#1{\textsf{#1}}\fi
\csname PreBibitemsHook\endcsname

\bibitem{huang2015community}
\begin{barticle}
\bauthor{\bsnm{Huang}, \binits{C.-C.}},
\bauthor{\bsnm{Lu}, \binits{Z.}}:
\batitle{Community challenges in biomedical text mining over 10 years: success,
  failure and the future}.
\bjtitle{Briefings in bioinformatics}
\bvolume{17}(\bissue{1}),
\bfpage{132}--\blpage{144}
(\byear{2015})
\end{barticle}
\endbibitem

\bibitem{demner2009can}
\begin{barticle}
\bauthor{\bsnm{Demner-Fushman}, \binits{D.}},
\bauthor{\bsnm{Chapman}, \binits{W.W.}},
\bauthor{\bsnm{McDonald}, \binits{C.J.}}:
\batitle{What can natural language processing do for clinical decision
  support?}
\bjtitle{Journal of biomedical informatics}
\bvolume{42}(\bissue{5}),
\bfpage{760}--\blpage{772}
(\byear{2009})
\end{barticle}
\endbibitem

\bibitem{wilcox2003role}
\begin{barticle}
\bauthor{\bsnm{Wilcox}, \binits{A.B.}},
\bauthor{\bsnm{Hripcsak}, \binits{G.}}:
\batitle{The role of domain knowledge in automating medical text report
  classification}.
\bjtitle{Journal of the American Medical Informatics Association}
\bvolume{10}(\bissue{4}),
\bfpage{330}--\blpage{338}
(\byear{2003})
\end{barticle}
\endbibitem

\bibitem{suominen2008machine}
\begin{bchapter}
\bauthor{\bsnm{Suominen}, \binits{H.}},
\bauthor{\bsnm{Ginter}, \binits{F.}},
\bauthor{\bsnm{Pyysalo}, \binits{S.}},
\bauthor{\bsnm{Airola}, \binits{A.}},
\bauthor{\bsnm{Pahikkala}, \binits{T.}},
\bauthor{\bsnm{Salanter}, \binits{S.}},
\bauthor{\bsnm{Salakoski}, \binits{T.}}:
\bctitle{Machine learning to automate the assignment of diagnosis codes to
  free-text radiology reports: a method description}.
In: \bbtitle{Proceedings of the ICML/UAI/COLT Workshop on Machine Learning for
  Health-Care Applications}
(\byear{2008})
\end{bchapter}
\endbibitem

\bibitem{solt2009semantic}
\begin{barticle}
\bauthor{\bsnm{Solt}, \binits{I.}},
\bauthor{\bsnm{Tikk}, \binits{D.}},
\bauthor{\bsnm{G{\'a}l}, \binits{V.}},
\bauthor{\bsnm{Kardkov{\'a}cs}, \binits{Z.T.}}:
\batitle{Semantic classification of diseases in discharge summaries using a
  context-aware rule-based classifier}.
\bjtitle{Journal of the American Medical Informatics Association}
\bvolume{16}(\bissue{4}),
\bfpage{580}--\blpage{584}
(\byear{2009})
\end{barticle}
\endbibitem

\bibitem{garla2013knowledge}
\begin{barticle}
\bauthor{\bsnm{Garla}, \binits{V.N.}},
\bauthor{\bsnm{Brandt}, \binits{C.}}:
\batitle{Knowledge-based biomedical word sense disambiguation: an evaluation
  and application to clinical document classification}.
\bjtitle{Journal of the American Medical Informatics Association}
\bvolume{20}(\bissue{5}),
\bfpage{882}--\blpage{886}
(\byear{2013})
\end{barticle}
\endbibitem

\bibitem{garla2012ontology}
\begin{barticle}
\bauthor{\bsnm{Garla}, \binits{V.N.}},
\bauthor{\bsnm{Brandt}, \binits{C.}}:
\batitle{Ontology-guided feature engineering for clinical text classification}.
\bjtitle{Journal of biomedical informatics}
\bvolume{45}(\bissue{5}),
\bfpage{992}--\blpage{998}
(\byear{2012})
\end{barticle}
\endbibitem

\bibitem{goodfellow2016deep}
\begin{bbook}
\bauthor{\bsnm{Goodfellow}, \binits{I.}},
\bauthor{\bsnm{Bengio}, \binits{Y.}},
\bauthor{\bsnm{Courville}, \binits{A.}},
\bauthor{\bsnm{Bengio}, \binits{Y.}}:
\bbtitle{Deep Learning}.
\bpublisher{MIT press Cambridge}, \blocation{???}
(\byear{2016})
\end{bbook}
\endbibitem

\bibitem{bodenreider2004unified}
\begin{barticle}
\bauthor{\bsnm{Bodenreider}, \binits{O.}}:
\batitle{The unified medical language system (umls): integrating biomedical
  terminology}.
\bjtitle{Nucleic acids research}
\bvolume{32}(\bissue{suppl\_1}),
\bfpage{267}--\blpage{270}
(\byear{2004})
\end{barticle}
\endbibitem

\bibitem{uzuner2008identifying}
\begin{barticle}
\bauthor{\bsnm{Uzuner}, \binits{{\"O}.}},
\bauthor{\bsnm{Goldstein}, \binits{I.}},
\bauthor{\bsnm{Luo}, \binits{Y.}},
\bauthor{\bsnm{Kohane}, \binits{I.}}:
\batitle{Identifying patient smoking status from medical discharge records}.
\bjtitle{Journal of the American Medical Informatics Association}
\bvolume{15}(\bissue{1}),
\bfpage{14}--\blpage{24}
(\byear{2008})
\end{barticle}
\endbibitem

\bibitem{stanfill2010systematic}
\begin{barticle}
\bauthor{\bsnm{Stanfill}, \binits{M.H.}},
\bauthor{\bsnm{Williams}, \binits{M.}},
\bauthor{\bsnm{Fenton}, \binits{S.H.}},
\bauthor{\bsnm{Jenders}, \binits{R.A.}},
\bauthor{\bsnm{Hersh}, \binits{W.R.}}:
\batitle{A systematic literature review of automated clinical coding and
  classification systems}.
\bjtitle{Journal of the American Medical Informatics Association}
\bvolume{17}(\bissue{6}),
\bfpage{646}--\blpage{651}
(\byear{2010})
\end{barticle}
\endbibitem

\bibitem{uzuner2009recognizing}
\begin{barticle}
\bauthor{\bsnm{Uzuner}, \binits{{\"O}.}}:
\batitle{Recognizing obesity and comorbidities in sparse data}.
\bjtitle{Journal of the American Medical Informatics Association}
\bvolume{16}(\bissue{4}),
\bfpage{561}--\blpage{570}
(\byear{2009})
\end{barticle}
\endbibitem

\bibitem{yao2016traditional}
\begin{bchapter}
\bauthor{\bsnm{Yao}, \binits{L.}},
\bauthor{\bsnm{Zhang}, \binits{Y.}},
\bauthor{\bsnm{Wei}, \binits{B.}},
\bauthor{\bsnm{Li}, \binits{Z.}},
\bauthor{\bsnm{Huang}, \binits{X.}}:
\bctitle{Traditional chinese medicine clinical records classification using
  knowledge-powered document embedding}.
In: \bbtitle{Bioinformatics and Biomedicine (BIBM), 2016 IEEE International
  Conference On},
pp. \bfpage{1926}--\blpage{1928}
(\byear{2016}).
\bcomment{IEEE}
\end{bchapter}
\endbibitem

\bibitem{bui2014learning}
\begin{barticle}
\bauthor{\bsnm{Bui}, \binits{D.D.A.}},
\bauthor{\bsnm{Zeng-Treitler}, \binits{Q.}}:
\batitle{Learning regular expressions for clinical text classification}.
\bjtitle{Journal of the American Medical Informatics Association}
\bvolume{21}(\bissue{5}),
\bfpage{850}--\blpage{857}
(\byear{2014})
\end{barticle}
\endbibitem

\bibitem{wang2010semi}
\begin{bchapter}
\bauthor{\bsnm{Wang}, \binits{Z.}},
\bauthor{\bsnm{Shawe-Taylor}, \binits{J.}},
\bauthor{\bsnm{Shah}, \binits{A.}}:
\bctitle{Semi-supervised feature learning from clinical text}.
In: \bbtitle{Bioinformatics and Biomedicine (BIBM), 2010 IEEE International
  Conference On},
pp. \bfpage{462}--\blpage{466}
(\byear{2010}).
\bcomment{IEEE}
\end{bchapter}
\endbibitem

\bibitem{garla2013semi}
\begin{barticle}
\bauthor{\bsnm{Garla}, \binits{V.}},
\bauthor{\bsnm{Taylor}, \binits{C.}},
\bauthor{\bsnm{Brandt}, \binits{C.}}:
\batitle{Semi-supervised clinical text classification with laplacian svms: an
  application to cancer case management}.
\bjtitle{Journal of biomedical informatics}
\bvolume{46}(\bissue{5}),
\bfpage{869}--\blpage{875}
(\byear{2013})
\end{barticle}
\endbibitem

\bibitem{figueroa2012active}
\begin{barticle}
\bauthor{\bsnm{Figueroa}, \binits{R.L.}},
\bauthor{\bsnm{Zeng-Treitler}, \binits{Q.}},
\bauthor{\bsnm{Ngo}, \binits{L.H.}},
\bauthor{\bsnm{Goryachev}, \binits{S.}},
\bauthor{\bsnm{Wiechmann}, \binits{E.P.}}:
\batitle{Active learning for clinical text classification: is it better than
  random sampling?}
\bjtitle{Journal of the American Medical Informatics Association}
\bvolume{19}(\bissue{5}),
\bfpage{809}--\blpage{816}
(\byear{2012})
\end{barticle}
\endbibitem

\bibitem{kim2014convolutional}
\begin{bchapter}
\bauthor{\bsnm{Kim}, \binits{Y.}}:
\bctitle{Convolutional neural networks for sentence classification}.
In: \bbtitle{Proceedings of the 2014 Conference on Empirical Methods in Natural
  Language Processing (EMNLP)},
pp. \bfpage{1746}--\blpage{1751}
(\byear{2014})
\end{bchapter}
\endbibitem

\bibitem{kalchbrenner2014convolutional}
\begin{bchapter}
\bauthor{\bsnm{Kalchbrenner}, \binits{N.}},
\bauthor{\bsnm{Grefenstette}, \binits{E.}},
\bauthor{\bsnm{Blunsom}, \binits{P.}}:
\bctitle{A convolutional neural network for modelling sentences}.
In: \bbtitle{Proceedings of the 52nd Annual Meeting of the Association for
  Computational Linguistics (Volume 1: Long Papers)},
vol. \bseriesno{1},
pp. \bfpage{655}--\blpage{665}
(\byear{2014})
\end{bchapter}
\endbibitem

\bibitem{tai2015improved}
\begin{bchapter}
\bauthor{\bsnm{Tai}, \binits{K.S.}},
\bauthor{\bsnm{Socher}, \binits{R.}},
\bauthor{\bsnm{Manning}, \binits{C.D.}}:
\bctitle{Improved semantic representations from tree-structured long short-term
  memory networks}.
In: \bbtitle{Proceedings of the 53rd Annual Meeting of the Association for
  Computational Linguistics and the 7th International Joint Conference on
  Natural Language Processing (Volume 1: Long Papers)},
vol. \bseriesno{1},
pp. \bfpage{1556}--\blpage{1566}
(\byear{2015})
\end{bchapter}
\endbibitem

\bibitem{yang2016hierarchical}
\begin{bchapter}
\bauthor{\bsnm{Yang}, \binits{Z.}},
\bauthor{\bsnm{Yang}, \binits{D.}},
\bauthor{\bsnm{Dyer}, \binits{C.}},
\bauthor{\bsnm{He}, \binits{X.}},
\bauthor{\bsnm{Smola}, \binits{A.}},
\bauthor{\bsnm{Hovy}, \binits{E.}}:
\bctitle{Hierarchical attention networks for document classification}.
In: \bbtitle{Proceedings of the 2016 Conference of the North American Chapter
  of the Association for Computational Linguistics: Human Language
  Technologies},
pp. \bfpage{1480}--\blpage{1489}
(\byear{2016})
\end{bchapter}
\endbibitem

\bibitem{beaulieu2016semi}
\begin{barticle}
\bauthor{\bsnm{Beaulieu-Jones}, \binits{B.K.}},
\bauthor{\bsnm{Greene}, \binits{C.S.}}, \betal:
\batitle{Semi-supervised learning of the electronic health record for phenotype
  stratification}.
\bjtitle{Journal of biomedical informatics}
\bvolume{64},
\bfpage{168}--\blpage{178}
(\byear{2016})
\end{barticle}
\endbibitem

\bibitem{gehrmann2017comparing}
\begin{botherref}
\oauthor{\bsnm{Gehrmann}, \binits{S.}},
\oauthor{\bsnm{Dernoncourt}, \binits{F.}},
\oauthor{\bsnm{Li}, \binits{Y.}},
\oauthor{\bsnm{Carlson}, \binits{E.T.}},
\oauthor{\bsnm{Wu}, \binits{J.T.}},
\oauthor{\bsnm{Welt}, \binits{J.}},
\oauthor{\bsnm{Foote~Jr}, \binits{J.}},
\oauthor{\bsnm{Moseley}, \binits{E.T.}},
\oauthor{\bsnm{Grant}, \binits{D.W.}},
\oauthor{\bsnm{Tyler}, \binits{P.D.}}, et al.:
Comparing rule-based and deep learning models for patient phenotyping.
arXiv preprint arXiv:1703.08705
(2017)
\end{botherref}
\endbibitem

\bibitem{uzuner20112010}
\begin{barticle}
\bauthor{\bsnm{Uzuner}, \binits{{\"O}.}},
\bauthor{\bsnm{South}, \binits{B.R.}},
\bauthor{\bsnm{Shen}, \binits{S.}},
\bauthor{\bsnm{DuVall}, \binits{S.L.}}:
\batitle{2010 i2b2/va challenge on concepts, assertions, and relations in
  clinical text}.
\bjtitle{Journal of the American Medical Informatics Association}
\bvolume{18}(\bissue{5}),
\bfpage{552}--\blpage{556}
(\byear{2011})
\end{barticle}
\endbibitem

\bibitem{luo2017recurrent}
\begin{barticle}
\bauthor{\bsnm{Luo}, \binits{Y.}}:
\batitle{Recurrent neural networks for classifying relations in clinical
  notes}.
\bjtitle{Journal of biomedical informatics}
\bvolume{72},
\bfpage{85}--\blpage{95}
(\byear{2017})
\end{barticle}
\endbibitem

\bibitem{luo2017segment}
\begin{barticle}
\bauthor{\bsnm{Luo}, \binits{Y.}},
\bauthor{\bsnm{Cheng}, \binits{Y.}},
\bauthor{\bsnm{Uzuner}, \binits{{\"O}.}},
\bauthor{\bsnm{Szolovits}, \binits{P.}},
\bauthor{\bsnm{Starren}, \binits{J.}}:
\batitle{Segment convolutional neural networks (seg-cnns) for classifying
  relations in clinical notes}.
\bjtitle{Journal of the American Medical Informatics Association}
\bvolume{25}(\bissue{1}),
\bfpage{93}--\blpage{98}
(\byear{2017})
\end{barticle}
\endbibitem

\bibitem{wu2015named}
\begin{barticle}
\bauthor{\bsnm{Wu}, \binits{Y.}},
\bauthor{\bsnm{Jiang}, \binits{M.}},
\bauthor{\bsnm{Lei}, \binits{J.}},
\bauthor{\bsnm{Xu}, \binits{H.}}:
\batitle{Named entity recognition in chinese clinical text using deep neural
  network}.
\bjtitle{Studies in health technology and informatics}
\bvolume{216},
\bfpage{624}
(\byear{2015})
\end{barticle}
\endbibitem

\bibitem{geraci2017applying}
\begin{barticle}
\bauthor{\bsnm{Geraci}, \binits{J.}},
\bauthor{\bsnm{Wilansky}, \binits{P.}},
\bauthor{\bparticle{de} \bsnm{Luca}, \binits{V.}},
\bauthor{\bsnm{Roy}, \binits{A.}},
\bauthor{\bsnm{Kennedy}, \binits{J.L.}},
\bauthor{\bsnm{Strauss}, \binits{J.}}:
\batitle{Applying deep neural networks to unstructured text notes in electronic
  medical records for phenotyping youth depression}.
\bjtitle{Evidence-based mental health}
\bvolume{20}(\bissue{3}),
\bfpage{83}--\blpage{87}
(\byear{2017})
\end{barticle}
\endbibitem

\bibitem{jagannatha2016structured}
\begin{bchapter}
\bauthor{\bsnm{Jagannatha}, \binits{A.N.}},
\bauthor{\bsnm{Yu}, \binits{H.}}:
\bctitle{Structured prediction models for rnn based sequence labeling in
  clinical text}.
In: \bbtitle{Proceedings of the Conference on Empirical Methods in Natural
  Language Processing. Conference on Empirical Methods in Natural Language
  Processing},
vol. \bseriesno{2016},
p. \bfpage{856}
(\byear{2016}).
\bcomment{NIH Public Access}
\end{bchapter}
\endbibitem

\bibitem{jagannatha2016bidirectional}
\begin{bchapter}
\bauthor{\bsnm{Jagannatha}, \binits{A.N.}},
\bauthor{\bsnm{Yu}, \binits{H.}}:
\bctitle{Bidirectional rnn for medical event detection in electronic health
  records}.
In: \bbtitle{Proceedings of the Conference. Association for Computational
  Linguistics. North American Chapter. Meeting},
vol. \bseriesno{2016},
p. \bfpage{473}
(\byear{2016}).
\bcomment{NIH Public Access}
\end{bchapter}
\endbibitem

\bibitem{lipton2015learning}
\begin{botherref}
\oauthor{\bsnm{Lipton}, \binits{Z.C.}},
\oauthor{\bsnm{Kale}, \binits{D.C.}},
\oauthor{\bsnm{Elkan}, \binits{C.}},
\oauthor{\bsnm{Wetzel}, \binits{R.}}:
Learning to diagnose with lstm recurrent neural networks.
arXiv preprint arXiv:1511.03677
(2015)
\end{botherref}
\endbibitem

\bibitem{che2015deep}
\begin{bchapter}
\bauthor{\bsnm{Che}, \binits{Z.}},
\bauthor{\bsnm{Kale}, \binits{D.}},
\bauthor{\bsnm{Li}, \binits{W.}},
\bauthor{\bsnm{Bahadori}, \binits{M.T.}},
\bauthor{\bsnm{Liu}, \binits{Y.}}:
\bctitle{Deep computational phenotyping}.
In: \bbtitle{Proceedings of the 21th ACM SIGKDD International Conference on
  Knowledge Discovery and Data Mining},
pp. \bfpage{507}--\blpage{516}
(\byear{2015}).
\bcomment{ACM}
\end{bchapter}
\endbibitem

\bibitem{mikolov2013distributed}
\begin{bchapter}
\bauthor{\bsnm{Mikolov}, \binits{T.}},
\bauthor{\bsnm{Sutskever}, \binits{I.}},
\bauthor{\bsnm{Chen}, \binits{K.}},
\bauthor{\bsnm{Corrado}, \binits{G.S.}},
\bauthor{\bsnm{Dean}, \binits{J.}}:
\bctitle{Distributed representations of words and phrases and their
  compositionality}.
In: \bbtitle{NIPS},
pp. \bfpage{3111}--\blpage{3119}
(\byear{2013})
\end{bchapter}
\endbibitem

\bibitem{johnson2016mimic}
\begin{barticle}
\bauthor{\bsnm{Johnson}, \binits{A.E.}},
\bauthor{\bsnm{Pollard}, \binits{T.J.}},
\bauthor{\bsnm{Shen}, \binits{L.}},
\bauthor{\bsnm{Li-wei}, \binits{H.L.}},
\bauthor{\bsnm{Feng}, \binits{M.}},
\bauthor{\bsnm{Ghassemi}, \binits{M.}},
\bauthor{\bsnm{Moody}, \binits{B.}},
\bauthor{\bsnm{Szolovits}, \binits{P.}},
\bauthor{\bsnm{Celi}, \binits{L.A.}},
\bauthor{\bsnm{Mark}, \binits{R.G.}}:
\batitle{Mimic-iii, a freely accessible critical care database}.
\bjtitle{Scientific data}
\bvolume{3},
\bfpage{160035}
(\byear{2016})
\end{barticle}
\endbibitem

\bibitem{aronson2010overview}
\begin{barticle}
\bauthor{\bsnm{Aronson}, \binits{A.R.}},
\bauthor{\bsnm{Lang}, \binits{F.-M.}}:
\batitle{An overview of metamap: historical perspective and recent advances}.
\bjtitle{Journal of the American Medical Informatics Association}
\bvolume{17}(\bissue{3}),
\bfpage{229}--\blpage{236}
(\byear{2010})
\end{barticle}
\endbibitem

\bibitem{de2014medical}
\begin{bchapter}
\bauthor{\bsnm{De~Vine}, \binits{L.}},
\bauthor{\bsnm{Zuccon}, \binits{G.}},
\bauthor{\bsnm{Koopman}, \binits{B.}},
\bauthor{\bsnm{Sitbon}, \binits{L.}},
\bauthor{\bsnm{Bruza}, \binits{P.}}:
\bctitle{Medical semantic similarity with a neural language model}.
In: \bbtitle{Proceedings of the 23rd ACM International Conference on Conference
  on Information and Knowledge Management},
pp. \bfpage{1819}--\blpage{1822}
(\byear{2014}).
\bcomment{ACM}
\end{bchapter}
\endbibitem

\bibitem{abadi2016tensorflow}
\begin{botherref}
\oauthor{\bsnm{Abadi}, \binits{M.}},
\oauthor{\bsnm{Barham}, \binits{P.}},
\oauthor{\bsnm{Chen}, \binits{J.}},
\oauthor{\bsnm{Chen}, \binits{Z.}},
\oauthor{\bsnm{Davis}, \binits{A.}},
\oauthor{\bsnm{Dean}, \binits{J.}},
\oauthor{\bsnm{Devin}, \binits{M.}},
\oauthor{\bsnm{Ghemawat}, \binits{S.}},
\oauthor{\bsnm{Irving}, \binits{G.}},
\oauthor{\bsnm{Isard}, \binits{M.}}, et al.:
Tensorflow: A system for large-scale machine learning.
\end{botherref}
\endbibitem

\bibitem{kinga2015method}
\begin{bchapter}
\bauthor{\bsnm{Kinga}, \binits{D.}},
\bauthor{\bsnm{Adam}, \binits{J.B.}}:
\bctitle{A method for stochastic optimization}.
In: \bbtitle{International Conference on Learning Representations (ICLR)}
(\byear{2015})
\end{bchapter}
\endbibitem

\bibitem{weng2017medical}
\begin{barticle}
\bauthor{\bsnm{Weng}, \binits{W.-H.}},
\bauthor{\bsnm{Wagholikar}, \binits{K.B.}},
\bauthor{\bsnm{McCray}, \binits{A.T.}},
\bauthor{\bsnm{Szolovits}, \binits{P.}},
\bauthor{\bsnm{Chueh}, \binits{H.C.}}:
\batitle{Medical subdomain classification of clinical notes using a machine
  learning-based natural language processing approach}.
\bjtitle{BMC medical informatics and decision making}
\bvolume{17}(\bissue{1}),
\bfpage{155}
(\byear{2017})
\end{barticle}
\endbibitem

\bibitem{zeng2017contralateral}
\begin{bchapter}
\bauthor{\bsnm{Zeng}, \binits{Z.}},
\bauthor{\bsnm{Li}, \binits{X.}},
\bauthor{\bsnm{Espino}, \binits{S.}},
\bauthor{\bsnm{Roy}, \binits{A.}},
\bauthor{\bsnm{Kitsch}, \binits{K.}},
\bauthor{\bsnm{Clare}, \binits{S.}},
\bauthor{\bsnm{Khan}, \binits{S.}},
\bauthor{\bsnm{Luo}, \binits{Y.}}:
\bctitle{Contralateral breast cancer event detection using nature language
  processing}.
In: \bbtitle{AMIA Annual Symposium Proceedings},
vol. \bseriesno{2017},
p. \bfpage{1885}
(\byear{2017}).
\bcomment{American Medical Informatics Association}
\end{bchapter}
\endbibitem

\end{thebibliography}

\newcommand{\BMCxmlcomment}[1]{}

\BMCxmlcomment{

<refgrp>

<bibl id="B1">
  <title><p>Community challenges in biomedical text mining over 10 years:
  success, failure and the future</p></title>
  <aug>
    <au><snm>Huang</snm><fnm>CC</fnm></au>
    <au><snm>Lu</snm><fnm>Z</fnm></au>
  </aug>
  <source>Briefings in bioinformatics</source>
  <publisher>Oxford University Press</publisher>
  <pubdate>2015</pubdate>
  <volume>17</volume>
  <issue>1</issue>
  <fpage>132</fpage>
  <lpage>-144</lpage>
</bibl>

<bibl id="B2">
  <title><p>What can natural language processing do for clinical decision
  support?</p></title>
  <aug>
    <au><snm>Demner Fushman</snm><fnm>D</fnm></au>
    <au><snm>Chapman</snm><fnm>WW</fnm></au>
    <au><snm>McDonald</snm><fnm>CJ</fnm></au>
  </aug>
  <source>Journal of biomedical informatics</source>
  <publisher>Elsevier</publisher>
  <pubdate>2009</pubdate>
  <volume>42</volume>
  <issue>5</issue>
  <fpage>760</fpage>
  <lpage>-772</lpage>
</bibl>

<bibl id="B3">
  <title><p>The role of domain knowledge in automating medical text report
  classification</p></title>
  <aug>
    <au><snm>Wilcox</snm><fnm>AB</fnm></au>
    <au><snm>Hripcsak</snm><fnm>G</fnm></au>
  </aug>
  <source>Journal of the American Medical Informatics Association</source>
  <publisher>Elsevier</publisher>
  <pubdate>2003</pubdate>
  <volume>10</volume>
  <issue>4</issue>
  <fpage>330</fpage>
  <lpage>-338</lpage>
</bibl>

<bibl id="B4">
  <title><p>Machine learning to automate the assignment of diagnosis codes to
  free-text radiology reports: a method description</p></title>
  <aug>
    <au><snm>Suominen</snm><fnm>H</fnm></au>
    <au><snm>Ginter</snm><fnm>F</fnm></au>
    <au><snm>Pyysalo</snm><fnm>S</fnm></au>
    <au><snm>Airola</snm><fnm>A</fnm></au>
    <au><snm>Pahikkala</snm><fnm>T</fnm></au>
    <au><snm>Salanter</snm><fnm>S</fnm></au>
    <au><snm>Salakoski</snm><fnm>T</fnm></au>
  </aug>
  <source>Proceedings of the ICML/UAI/COLT Workshop on Machine Learning for
  Health-Care Applications</source>
  <pubdate>2008</pubdate>
</bibl>

<bibl id="B5">
  <title><p>Semantic classification of diseases in discharge summaries using a
  context-aware rule-based classifier</p></title>
  <aug>
    <au><snm>Solt</snm><fnm>I</fnm></au>
    <au><snm>Tikk</snm><fnm>D</fnm></au>
    <au><snm>G{\'a}l</snm><fnm>V</fnm></au>
    <au><snm>Kardkov{\'a}cs</snm><fnm>ZT</fnm></au>
  </aug>
  <source>Journal of the American Medical Informatics Association</source>
  <publisher>BMJ Group BMA House, Tavistock Square, London, WC1H
  9JR</publisher>
  <pubdate>2009</pubdate>
  <volume>16</volume>
  <issue>4</issue>
  <fpage>580</fpage>
  <lpage>-584</lpage>
</bibl>

<bibl id="B6">
  <title><p>Knowledge-based biomedical word sense disambiguation: an evaluation
  and application to clinical document classification</p></title>
  <aug>
    <au><snm>Garla</snm><fnm>VN</fnm></au>
    <au><snm>Brandt</snm><fnm>C</fnm></au>
  </aug>
  <source>Journal of the American Medical Informatics Association</source>
  <publisher>The Oxford University Press</publisher>
  <pubdate>2013</pubdate>
  <volume>20</volume>
  <issue>5</issue>
  <fpage>882</fpage>
  <lpage>-886</lpage>
</bibl>

<bibl id="B7">
  <title><p>Ontology-guided feature engineering for clinical text
  classification</p></title>
  <aug>
    <au><snm>Garla</snm><fnm>VN</fnm></au>
    <au><snm>Brandt</snm><fnm>C</fnm></au>
  </aug>
  <source>Journal of biomedical informatics</source>
  <publisher>Elsevier</publisher>
  <pubdate>2012</pubdate>
  <volume>45</volume>
  <issue>5</issue>
  <fpage>992</fpage>
  <lpage>-998</lpage>
</bibl>

<bibl id="B8">
  <title><p>Deep learning</p></title>
  <aug>
    <au><snm>Goodfellow</snm><fnm>I</fnm></au>
    <au><snm>Bengio</snm><fnm>Y</fnm></au>
    <au><snm>Courville</snm><fnm>A</fnm></au>
    <au><snm>Bengio</snm><fnm>Y</fnm></au>
  </aug>
  <publisher>MIT press Cambridge</publisher>
  <pubdate>2016</pubdate>
</bibl>

<bibl id="B9">
  <title><p>The unified medical language system (UMLS): integrating biomedical
  terminology</p></title>
  <aug>
    <au><snm>Bodenreider</snm><fnm>O</fnm></au>
  </aug>
  <source>Nucleic acids research</source>
  <publisher>Oxford University Press</publisher>
  <pubdate>2004</pubdate>
  <volume>32</volume>
  <issue>suppl\_1</issue>
  <fpage>D267</fpage>
  <lpage>-D270</lpage>
</bibl>

<bibl id="B10">
  <title><p>Identifying patient smoking status from medical discharge
  records</p></title>
  <aug>
    <au><snm>Uzuner</snm><fnm>{\"O}</fnm></au>
    <au><snm>Goldstein</snm><fnm>I</fnm></au>
    <au><snm>Luo</snm><fnm>Y</fnm></au>
    <au><snm>Kohane</snm><fnm>I</fnm></au>
  </aug>
  <source>Journal of the American Medical Informatics Association</source>
  <publisher>Elsevier</publisher>
  <pubdate>2008</pubdate>
  <volume>15</volume>
  <issue>1</issue>
  <fpage>14</fpage>
  <lpage>-24</lpage>
</bibl>

<bibl id="B11">
  <title><p>A systematic literature review of automated clinical coding and
  classification systems</p></title>
  <aug>
    <au><snm>Stanfill</snm><fnm>MH</fnm></au>
    <au><snm>Williams</snm><fnm>M</fnm></au>
    <au><snm>Fenton</snm><fnm>SH</fnm></au>
    <au><snm>Jenders</snm><fnm>RA</fnm></au>
    <au><snm>Hersh</snm><fnm>WR</fnm></au>
  </aug>
  <source>Journal of the American Medical Informatics Association</source>
  <publisher>The Oxford University Press</publisher>
  <pubdate>2010</pubdate>
  <volume>17</volume>
  <issue>6</issue>
  <fpage>646</fpage>
  <lpage>-651</lpage>
</bibl>

<bibl id="B12">
  <title><p>Recognizing obesity and comorbidities in sparse data</p></title>
  <aug>
    <au><snm>Uzuner</snm><fnm>{\"O}</fnm></au>
  </aug>
  <source>Journal of the American Medical Informatics Association</source>
  <publisher>The Oxford University Press</publisher>
  <pubdate>2009</pubdate>
  <volume>16</volume>
  <issue>4</issue>
  <fpage>561</fpage>
  <lpage>-570</lpage>
</bibl>

<bibl id="B13">
  <title><p>Traditional Chinese medicine clinical records classification using
  knowledge-powered document embedding</p></title>
  <aug>
    <au><snm>Yao</snm><fnm>L</fnm></au>
    <au><snm>Zhang</snm><fnm>Y</fnm></au>
    <au><snm>Wei</snm><fnm>B</fnm></au>
    <au><snm>Li</snm><fnm>Z</fnm></au>
    <au><snm>Huang</snm><fnm>X</fnm></au>
  </aug>
  <source>Bioinformatics and Biomedicine (BIBM), 2016 IEEE International
  Conference on</source>
  <pubdate>2016</pubdate>
  <fpage>1926</fpage>
  <lpage>-1928</lpage>
</bibl>

<bibl id="B14">
  <title><p>Learning regular expressions for clinical text
  classification</p></title>
  <aug>
    <au><snm>Bui</snm><fnm>DDA</fnm></au>
    <au><snm>Zeng Treitler</snm><fnm>Q</fnm></au>
  </aug>
  <source>Journal of the American Medical Informatics Association</source>
  <publisher>The Oxford University Press</publisher>
  <pubdate>2014</pubdate>
  <volume>21</volume>
  <issue>5</issue>
  <fpage>850</fpage>
  <lpage>-857</lpage>
</bibl>

<bibl id="B15">
  <title><p>Semi-supervised feature learning from clinical text</p></title>
  <aug>
    <au><snm>Wang</snm><fnm>Z</fnm></au>
    <au><snm>Shawe Taylor</snm><fnm>J</fnm></au>
    <au><snm>Shah</snm><fnm>A</fnm></au>
  </aug>
  <source>Bioinformatics and Biomedicine (BIBM), 2010 IEEE International
  Conference on</source>
  <pubdate>2010</pubdate>
  <fpage>462</fpage>
  <lpage>-466</lpage>
</bibl>

<bibl id="B16">
  <title><p>Semi-supervised clinical text classification with Laplacian SVMs:
  an application to cancer case management</p></title>
  <aug>
    <au><snm>Garla</snm><fnm>V</fnm></au>
    <au><snm>Taylor</snm><fnm>C</fnm></au>
    <au><snm>Brandt</snm><fnm>C</fnm></au>
  </aug>
  <source>Journal of biomedical informatics</source>
  <publisher>Elsevier</publisher>
  <pubdate>2013</pubdate>
  <volume>46</volume>
  <issue>5</issue>
  <fpage>869</fpage>
  <lpage>-875</lpage>
</bibl>

<bibl id="B17">
  <title><p>Active learning for clinical text classification: is it better than
  random sampling?</p></title>
  <aug>
    <au><snm>Figueroa</snm><fnm>RL</fnm></au>
    <au><snm>Zeng Treitler</snm><fnm>Q</fnm></au>
    <au><snm>Ngo</snm><fnm>LH</fnm></au>
    <au><snm>Goryachev</snm><fnm>S</fnm></au>
    <au><snm>Wiechmann</snm><fnm>EP</fnm></au>
  </aug>
  <source>Journal of the American Medical Informatics Association</source>
  <publisher>The Oxford University Press</publisher>
  <pubdate>2012</pubdate>
  <volume>19</volume>
  <issue>5</issue>
  <fpage>809</fpage>
  <lpage>-816</lpage>
</bibl>

<bibl id="B18">
  <title><p>Convolutional Neural Networks for Sentence
  Classification</p></title>
  <aug>
    <au><snm>Kim</snm><fnm>Y</fnm></au>
  </aug>
  <source>Proceedings of the 2014 Conference on Empirical Methods in Natural
  Language Processing (EMNLP)</source>
  <pubdate>2014</pubdate>
  <fpage>1746</fpage>
  <lpage>-1751</lpage>
</bibl>

<bibl id="B19">
  <title><p>A Convolutional Neural Network for Modelling Sentences</p></title>
  <aug>
    <au><snm>Kalchbrenner</snm><fnm>N</fnm></au>
    <au><snm>Grefenstette</snm><fnm>E</fnm></au>
    <au><snm>Blunsom</snm><fnm>P</fnm></au>
  </aug>
  <source>Proceedings of the 52nd Annual Meeting of the Association for
  Computational Linguistics (Volume 1: Long Papers)</source>
  <pubdate>2014</pubdate>
  <volume>1</volume>
  <fpage>655</fpage>
  <lpage>-665</lpage>
</bibl>

<bibl id="B20">
  <title><p>Improved Semantic Representations From Tree-Structured Long
  Short-Term Memory Networks</p></title>
  <aug>
    <au><snm>Tai</snm><fnm>KS</fnm></au>
    <au><snm>Socher</snm><fnm>R</fnm></au>
    <au><snm>Manning</snm><fnm>CD</fnm></au>
  </aug>
  <source>Proceedings of the 53rd Annual Meeting of the Association for
  Computational Linguistics and the 7th International Joint Conference on
  Natural Language Processing (Volume 1: Long Papers)</source>
  <pubdate>2015</pubdate>
  <volume>1</volume>
  <fpage>1556</fpage>
  <lpage>-1566</lpage>
</bibl>

<bibl id="B21">
  <title><p>Hierarchical attention networks for document
  classification</p></title>
  <aug>
    <au><snm>Yang</snm><fnm>Z</fnm></au>
    <au><snm>Yang</snm><fnm>D</fnm></au>
    <au><snm>Dyer</snm><fnm>C</fnm></au>
    <au><snm>He</snm><fnm>X</fnm></au>
    <au><snm>Smola</snm><fnm>A</fnm></au>
    <au><snm>Hovy</snm><fnm>E</fnm></au>
  </aug>
  <source>Proceedings of the 2016 Conference of the North American Chapter of
  the Association for Computational Linguistics: Human Language
  Technologies</source>
  <pubdate>2016</pubdate>
  <fpage>1480</fpage>
  <lpage>-1489</lpage>
</bibl>

<bibl id="B22">
  <title><p>Semi-supervised learning of the electronic health record for
  phenotype stratification</p></title>
  <aug>
    <au><snm>Beaulieu Jones</snm><fnm>BK</fnm></au>
    <au><snm>Greene</snm><fnm>CS</fnm></au>
    <au><cnm>others</cnm></au>
  </aug>
  <source>Journal of biomedical informatics</source>
  <publisher>Elsevier</publisher>
  <pubdate>2016</pubdate>
  <volume>64</volume>
  <fpage>168</fpage>
  <lpage>-178</lpage>
</bibl>

<bibl id="B23">
  <title><p>Comparing Rule-Based and Deep Learning Models for Patient
  Phenotyping</p></title>
  <aug>
    <au><snm>Gehrmann</snm><fnm>S</fnm></au>
    <au><snm>Dernoncourt</snm><fnm>F</fnm></au>
    <au><snm>Li</snm><fnm>Y</fnm></au>
    <au><snm>Carlson</snm><fnm>ET</fnm></au>
    <au><snm>Wu</snm><fnm>JT</fnm></au>
    <au><snm>Welt</snm><fnm>J</fnm></au>
    <au><snm>Foote Jr</snm><fnm>J</fnm></au>
    <au><snm>Moseley</snm><fnm>ET</fnm></au>
    <au><snm>Grant</snm><fnm>DW</fnm></au>
    <au><snm>Tyler</snm><fnm>PD</fnm></au>
    <au><cnm>others</cnm></au>
  </aug>
  <source>arXiv preprint arXiv:1703.08705</source>
  <pubdate>2017</pubdate>
</bibl>

<bibl id="B24">
  <title><p>2010 i2b2/VA challenge on concepts, assertions, and relations in
  clinical text</p></title>
  <aug>
    <au><snm>Uzuner</snm><fnm>{\"O}</fnm></au>
    <au><snm>South</snm><fnm>BR</fnm></au>
    <au><snm>Shen</snm><fnm>S</fnm></au>
    <au><snm>DuVall</snm><fnm>SL</fnm></au>
  </aug>
  <source>Journal of the American Medical Informatics Association</source>
  <publisher>BMJ Group BMA House, Tavistock Square, London, WC1H
  9JR</publisher>
  <pubdate>2011</pubdate>
  <volume>18</volume>
  <issue>5</issue>
  <fpage>552</fpage>
  <lpage>-556</lpage>
</bibl>

<bibl id="B25">
  <title><p>Recurrent neural networks for classifying relations in clinical
  notes</p></title>
  <aug>
    <au><snm>Luo</snm><fnm>Y</fnm></au>
  </aug>
  <source>Journal of biomedical informatics</source>
  <publisher>Elsevier</publisher>
  <pubdate>2017</pubdate>
  <volume>72</volume>
  <fpage>85</fpage>
  <lpage>-95</lpage>
</bibl>

<bibl id="B26">
  <title><p>Segment convolutional neural networks (Seg-CNNs) for classifying
  relations in clinical notes</p></title>
  <aug>
    <au><snm>Luo</snm><fnm>Y</fnm></au>
    <au><snm>Cheng</snm><fnm>Y</fnm></au>
    <au><snm>Uzuner</snm><fnm>{\"O}</fnm></au>
    <au><snm>Szolovits</snm><fnm>P</fnm></au>
    <au><snm>Starren</snm><fnm>J</fnm></au>
  </aug>
  <source>Journal of the American Medical Informatics Association</source>
  <publisher>Oxford University Press</publisher>
  <pubdate>2017</pubdate>
  <volume>25</volume>
  <issue>1</issue>
  <fpage>93</fpage>
  <lpage>-98</lpage>
</bibl>

<bibl id="B27">
  <title><p>Named entity recognition in Chinese clinical text using deep neural
  network</p></title>
  <aug>
    <au><snm>Wu</snm><fnm>Y</fnm></au>
    <au><snm>Jiang</snm><fnm>M</fnm></au>
    <au><snm>Lei</snm><fnm>J</fnm></au>
    <au><snm>Xu</snm><fnm>H</fnm></au>
  </aug>
  <source>Studies in health technology and informatics</source>
  <publisher>NIH Public Access</publisher>
  <pubdate>2015</pubdate>
  <volume>216</volume>
  <fpage>624</fpage>
</bibl>

<bibl id="B28">
  <title><p>Applying deep neural networks to unstructured text notes in
  electronic medical records for phenotyping youth depression</p></title>
  <aug>
    <au><snm>Geraci</snm><fnm>J</fnm></au>
    <au><snm>Wilansky</snm><fnm>P</fnm></au>
    <au><snm>Luca</snm><fnm>V</fnm></au>
    <au><snm>Roy</snm><fnm>A</fnm></au>
    <au><snm>Kennedy</snm><fnm>JL</fnm></au>
    <au><snm>Strauss</snm><fnm>J</fnm></au>
  </aug>
  <source>Evidence-based mental health</source>
  <publisher>Royal College of Psychiatrists</publisher>
  <pubdate>2017</pubdate>
  <volume>20</volume>
  <issue>3</issue>
  <fpage>83</fpage>
  <lpage>-87</lpage>
</bibl>

<bibl id="B29">
  <title><p>Structured prediction models for RNN based sequence labeling in
  clinical text</p></title>
  <aug>
    <au><snm>Jagannatha</snm><fnm>AN</fnm></au>
    <au><snm>Yu</snm><fnm>H</fnm></au>
  </aug>
  <source>Proceedings of the Conference on Empirical Methods in Natural
  Language Processing. Conference on Empirical Methods in Natural Language
  Processing</source>
  <pubdate>2016</pubdate>
  <volume>2016</volume>
  <fpage>856</fpage>
</bibl>

<bibl id="B30">
  <title><p>Bidirectional RNN for medical event detection in electronic health
  records</p></title>
  <aug>
    <au><snm>Jagannatha</snm><fnm>AN</fnm></au>
    <au><snm>Yu</snm><fnm>H</fnm></au>
  </aug>
  <source>Proceedings of the conference. Association for Computational
  Linguistics. North American Chapter. Meeting</source>
  <pubdate>2016</pubdate>
  <volume>2016</volume>
  <fpage>473</fpage>
</bibl>

<bibl id="B31">
  <title><p>Learning to diagnose with LSTM recurrent neural
  networks</p></title>
  <aug>
    <au><snm>Lipton</snm><fnm>ZC</fnm></au>
    <au><snm>Kale</snm><fnm>DC</fnm></au>
    <au><snm>Elkan</snm><fnm>C</fnm></au>
    <au><snm>Wetzel</snm><fnm>R</fnm></au>
  </aug>
  <source>arXiv preprint arXiv:1511.03677</source>
  <pubdate>2015</pubdate>
</bibl>

<bibl id="B32">
  <title><p>Deep computational phenotyping</p></title>
  <aug>
    <au><snm>Che</snm><fnm>Z</fnm></au>
    <au><snm>Kale</snm><fnm>D</fnm></au>
    <au><snm>Li</snm><fnm>W</fnm></au>
    <au><snm>Bahadori</snm><fnm>MT</fnm></au>
    <au><snm>Liu</snm><fnm>Y</fnm></au>
  </aug>
  <source>Proceedings of the 21th ACM SIGKDD International Conference on
  Knowledge Discovery and Data Mining</source>
  <pubdate>2015</pubdate>
  <fpage>507</fpage>
  <lpage>-516</lpage>
</bibl>

<bibl id="B33">
  <title><p>Distributed representations of words and phrases and their
  compositionality</p></title>
  <aug>
    <au><snm>Mikolov</snm><fnm>T</fnm></au>
    <au><snm>Sutskever</snm><fnm>I</fnm></au>
    <au><snm>Chen</snm><fnm>K</fnm></au>
    <au><snm>Corrado</snm><fnm>GS</fnm></au>
    <au><snm>Dean</snm><fnm>J</fnm></au>
  </aug>
  <source>NIPS</source>
  <pubdate>2013</pubdate>
  <fpage>3111</fpage>
  <lpage>-3119</lpage>
</bibl>

<bibl id="B34">
  <title><p>MIMIC-III, a freely accessible critical care database</p></title>
  <aug>
    <au><snm>Johnson</snm><fnm>AE</fnm></au>
    <au><snm>Pollard</snm><fnm>TJ</fnm></au>
    <au><snm>Shen</snm><fnm>L</fnm></au>
    <au><snm>Li wei</snm><fnm>HL</fnm></au>
    <au><snm>Feng</snm><fnm>M</fnm></au>
    <au><snm>Ghassemi</snm><fnm>M</fnm></au>
    <au><snm>Moody</snm><fnm>B</fnm></au>
    <au><snm>Szolovits</snm><fnm>P</fnm></au>
    <au><snm>Celi</snm><fnm>LA</fnm></au>
    <au><snm>Mark</snm><fnm>RG</fnm></au>
  </aug>
  <source>Scientific data</source>
  <publisher>Nature Publishing Group</publisher>
  <pubdate>2016</pubdate>
  <volume>3</volume>
  <fpage>160035</fpage>
</bibl>

<bibl id="B35">
  <title><p>An overview of MetaMap: historical perspective and recent
  advances</p></title>
  <aug>
    <au><snm>Aronson</snm><fnm>AR</fnm></au>
    <au><snm>Lang</snm><fnm>FM</fnm></au>
  </aug>
  <source>Journal of the American Medical Informatics Association</source>
  <publisher>BMJ Group BMA House, Tavistock Square, London, WC1H
  9JR</publisher>
  <pubdate>2010</pubdate>
  <volume>17</volume>
  <issue>3</issue>
  <fpage>229</fpage>
  <lpage>-236</lpage>
</bibl>

<bibl id="B36">
  <title><p>Medical semantic similarity with a neural language
  model</p></title>
  <aug>
    <au><snm>De Vine</snm><fnm>L</fnm></au>
    <au><snm>Zuccon</snm><fnm>G</fnm></au>
    <au><snm>Koopman</snm><fnm>B</fnm></au>
    <au><snm>Sitbon</snm><fnm>L</fnm></au>
    <au><snm>Bruza</snm><fnm>P</fnm></au>
  </aug>
  <source>Proceedings of the 23rd ACM international conference on conference on
  information and knowledge management</source>
  <pubdate>2014</pubdate>
  <fpage>1819</fpage>
  <lpage>-1822</lpage>
</bibl>

<bibl id="B37">
  <title><p>TensorFlow: A System for Large-Scale Machine Learning.</p></title>
  <aug>
    <au><snm>Abadi</snm><fnm>M</fnm></au>
    <au><snm>Barham</snm><fnm>P</fnm></au>
    <au><snm>Chen</snm><fnm>J</fnm></au>
    <au><snm>Chen</snm><fnm>Z</fnm></au>
    <au><snm>Davis</snm><fnm>A</fnm></au>
    <au><snm>Dean</snm><fnm>J</fnm></au>
    <au><snm>Devin</snm><fnm>M</fnm></au>
    <au><snm>Ghemawat</snm><fnm>S</fnm></au>
    <au><snm>Irving</snm><fnm>G</fnm></au>
    <au><snm>Isard</snm><fnm>M</fnm></au>
    <au><cnm>others</cnm></au>
  </aug>
</bibl>

<bibl id="B38">
  <title><p>A method for stochastic optimization</p></title>
  <aug>
    <au><snm>Kinga</snm><fnm>D</fnm></au>
    <au><snm>Adam</snm><fnm>JB</fnm></au>
  </aug>
  <source>International Conference on Learning Representations (ICLR)</source>
  <pubdate>2015</pubdate>
</bibl>

<bibl id="B39">
  <title><p>Medical subdomain classification of clinical notes using a machine
  learning-based natural language processing approach</p></title>
  <aug>
    <au><snm>Weng</snm><fnm>WH</fnm></au>
    <au><snm>Wagholikar</snm><fnm>KB</fnm></au>
    <au><snm>McCray</snm><fnm>AT</fnm></au>
    <au><snm>Szolovits</snm><fnm>P</fnm></au>
    <au><snm>Chueh</snm><fnm>HC</fnm></au>
  </aug>
  <source>BMC medical informatics and decision making</source>
  <publisher>BioMed Central</publisher>
  <pubdate>2017</pubdate>
  <volume>17</volume>
  <issue>1</issue>
  <fpage>155</fpage>
</bibl>

<bibl id="B40">
  <title><p>Contralateral Breast Cancer Event Detection Using Nature Language
  Processing</p></title>
  <aug>
    <au><snm>Zeng</snm><fnm>Z</fnm></au>
    <au><snm>Li</snm><fnm>X</fnm></au>
    <au><snm>Espino</snm><fnm>S</fnm></au>
    <au><snm>Roy</snm><fnm>A</fnm></au>
    <au><snm>Kitsch</snm><fnm>K</fnm></au>
    <au><snm>Clare</snm><fnm>S</fnm></au>
    <au><snm>Khan</snm><fnm>S</fnm></au>
    <au><snm>Luo</snm><fnm>Y</fnm></au>
  </aug>
  <source>AMIA Annual Symposium Proceedings</source>
  <pubdate>2017</pubdate>
  <volume>2017</volume>
  <fpage>1885</fpage>
</bibl>

</refgrp>
} 





\newcommand{\BMCxmlcomment}[1]{}

\BMCxmlcomment{

<refgrp>

<bibl id="B1">
  <title><p>Community challenges in biomedical text mining over 10 years:
  success, failure and the future</p></title>
  <aug>
    <au><snm>Huang</snm><fnm>CC</fnm></au>
    <au><snm>Lu</snm><fnm>Z</fnm></au>
  </aug>
  <source>Briefings in bioinformatics</source>
  <publisher>Oxford University Press</publisher>
  <pubdate>2015</pubdate>
  <volume>17</volume>
  <issue>1</issue>
  <fpage>132</fpage>
  <lpage>-144</lpage>
</bibl>

<bibl id="B2">
  <title><p>What can natural language processing do for clinical decision
  support?</p></title>
  <aug>
    <au><snm>Demner Fushman</snm><fnm>D</fnm></au>
    <au><snm>Chapman</snm><fnm>WW</fnm></au>
    <au><snm>McDonald</snm><fnm>CJ</fnm></au>
  </aug>
  <source>Journal of biomedical informatics</source>
  <publisher>Elsevier</publisher>
  <pubdate>2009</pubdate>
  <volume>42</volume>
  <issue>5</issue>
  <fpage>760</fpage>
  <lpage>-772</lpage>
</bibl>

<bibl id="B3">
  <title><p>The role of domain knowledge in automating medical text report
  classification</p></title>
  <aug>
    <au><snm>Wilcox</snm><fnm>AB</fnm></au>
    <au><snm>Hripcsak</snm><fnm>G</fnm></au>
  </aug>
  <source>Journal of the American Medical Informatics Association</source>
  <publisher>Elsevier</publisher>
  <pubdate>2003</pubdate>
  <volume>10</volume>
  <issue>4</issue>
  <fpage>330</fpage>
  <lpage>-338</lpage>
</bibl>

<bibl id="B4">
  <title><p>Machine learning to automate the assignment of diagnosis codes to
  free-text radiology reports: a method description</p></title>
  <aug>
    <au><snm>Suominen</snm><fnm>H</fnm></au>
    <au><snm>Ginter</snm><fnm>F</fnm></au>
    <au><snm>Pyysalo</snm><fnm>S</fnm></au>
    <au><snm>Airola</snm><fnm>A</fnm></au>
    <au><snm>Pahikkala</snm><fnm>T</fnm></au>
    <au><snm>Salanter</snm><fnm>S</fnm></au>
    <au><snm>Salakoski</snm><fnm>T</fnm></au>
  </aug>
  <source>Proceedings of the ICML/UAI/COLT Workshop on Machine Learning for
  Health-Care Applications</source>
  <pubdate>2008</pubdate>
</bibl>

<bibl id="B5">
  <title><p>Semantic classification of diseases in discharge summaries using a
  context-aware rule-based classifier</p></title>
  <aug>
    <au><snm>Solt</snm><fnm>I</fnm></au>
    <au><snm>Tikk</snm><fnm>D</fnm></au>
    <au><snm>G{\'a}l</snm><fnm>V</fnm></au>
    <au><snm>Kardkov{\'a}cs</snm><fnm>ZT</fnm></au>
  </aug>
  <source>Journal of the American Medical Informatics Association</source>
  <publisher>BMJ Group BMA House, Tavistock Square, London, WC1H
  9JR</publisher>
  <pubdate>2009</pubdate>
  <volume>16</volume>
  <issue>4</issue>
  <fpage>580</fpage>
  <lpage>-584</lpage>
</bibl>

<bibl id="B6">
  <title><p>Knowledge-based biomedical word sense disambiguation: an evaluation
  and application to clinical document classification</p></title>
  <aug>
    <au><snm>Garla</snm><fnm>VN</fnm></au>
    <au><snm>Brandt</snm><fnm>C</fnm></au>
  </aug>
  <source>Journal of the American Medical Informatics Association</source>
  <publisher>The Oxford University Press</publisher>
  <pubdate>2013</pubdate>
  <volume>20</volume>
  <issue>5</issue>
  <fpage>882</fpage>
  <lpage>-886</lpage>
</bibl>

<bibl id="B7">
  <title><p>Ontology-guided feature engineering for clinical text
  classification</p></title>
  <aug>
    <au><snm>Garla</snm><fnm>VN</fnm></au>
    <au><snm>Brandt</snm><fnm>C</fnm></au>
  </aug>
  <source>Journal of biomedical informatics</source>
  <publisher>Elsevier</publisher>
  <pubdate>2012</pubdate>
  <volume>45</volume>
  <issue>5</issue>
  <fpage>992</fpage>
  <lpage>-998</lpage>
</bibl>

<bibl id="B8">
  <title><p>Deep learning</p></title>
  <aug>
    <au><snm>Goodfellow</snm><fnm>I</fnm></au>
    <au><snm>Bengio</snm><fnm>Y</fnm></au>
    <au><snm>Courville</snm><fnm>A</fnm></au>
    <au><snm>Bengio</snm><fnm>Y</fnm></au>
  </aug>
  <publisher>MIT press Cambridge</publisher>
  <pubdate>2016</pubdate>
</bibl>

<bibl id="B9">
  <title><p>The unified medical language system (UMLS): integrating biomedical
  terminology</p></title>
  <aug>
    <au><snm>Bodenreider</snm><fnm>O</fnm></au>
  </aug>
  <source>Nucleic acids research</source>
  <publisher>Oxford University Press</publisher>
  <pubdate>2004</pubdate>
  <volume>32</volume>
  <issue>suppl\_1</issue>
  <fpage>D267</fpage>
  <lpage>-D270</lpage>
</bibl>

<bibl id="B10">
  <title><p>Identifying patient smoking status from medical discharge
  records</p></title>
  <aug>
    <au><snm>Uzuner</snm><fnm>{\"O}</fnm></au>
    <au><snm>Goldstein</snm><fnm>I</fnm></au>
    <au><snm>Luo</snm><fnm>Y</fnm></au>
    <au><snm>Kohane</snm><fnm>I</fnm></au>
  </aug>
  <source>Journal of the American Medical Informatics Association</source>
  <publisher>Elsevier</publisher>
  <pubdate>2008</pubdate>
  <volume>15</volume>
  <issue>1</issue>
  <fpage>14</fpage>
  <lpage>-24</lpage>
</bibl>

<bibl id="B11">
  <title><p>A systematic literature review of automated clinical coding and
  classification systems</p></title>
  <aug>
    <au><snm>Stanfill</snm><fnm>MH</fnm></au>
    <au><snm>Williams</snm><fnm>M</fnm></au>
    <au><snm>Fenton</snm><fnm>SH</fnm></au>
    <au><snm>Jenders</snm><fnm>RA</fnm></au>
    <au><snm>Hersh</snm><fnm>WR</fnm></au>
  </aug>
  <source>Journal of the American Medical Informatics Association</source>
  <publisher>The Oxford University Press</publisher>
  <pubdate>2010</pubdate>
  <volume>17</volume>
  <issue>6</issue>
  <fpage>646</fpage>
  <lpage>-651</lpage>
</bibl>

<bibl id="B12">
  <title><p>Recognizing obesity and comorbidities in sparse data</p></title>
  <aug>
    <au><snm>Uzuner</snm><fnm>{\"O}</fnm></au>
  </aug>
  <source>Journal of the American Medical Informatics Association</source>
  <publisher>The Oxford University Press</publisher>
  <pubdate>2009</pubdate>
  <volume>16</volume>
  <issue>4</issue>
  <fpage>561</fpage>
  <lpage>-570</lpage>
</bibl>

<bibl id="B13">
  <title><p>Traditional Chinese medicine clinical records classification using
  knowledge-powered document embedding</p></title>
  <aug>
    <au><snm>Yao</snm><fnm>L</fnm></au>
    <au><snm>Zhang</snm><fnm>Y</fnm></au>
    <au><snm>Wei</snm><fnm>B</fnm></au>
    <au><snm>Li</snm><fnm>Z</fnm></au>
    <au><snm>Huang</snm><fnm>X</fnm></au>
  </aug>
  <source>Bioinformatics and Biomedicine (BIBM), 2016 IEEE International
  Conference on</source>
  <pubdate>2016</pubdate>
  <fpage>1926</fpage>
  <lpage>-1928</lpage>
</bibl>

<bibl id="B14">
  <title><p>Learning regular expressions for clinical text
  classification</p></title>
  <aug>
    <au><snm>Bui</snm><fnm>DDA</fnm></au>
    <au><snm>Zeng Treitler</snm><fnm>Q</fnm></au>
  </aug>
  <source>Journal of the American Medical Informatics Association</source>
  <publisher>The Oxford University Press</publisher>
  <pubdate>2014</pubdate>
  <volume>21</volume>
  <issue>5</issue>
  <fpage>850</fpage>
  <lpage>-857</lpage>
</bibl>

<bibl id="B15">
  <title><p>Semi-supervised feature learning from clinical text</p></title>
  <aug>
    <au><snm>Wang</snm><fnm>Z</fnm></au>
    <au><snm>Shawe Taylor</snm><fnm>J</fnm></au>
    <au><snm>Shah</snm><fnm>A</fnm></au>
  </aug>
  <source>Bioinformatics and Biomedicine (BIBM), 2010 IEEE International
  Conference on</source>
  <pubdate>2010</pubdate>
  <fpage>462</fpage>
  <lpage>-466</lpage>
</bibl>

<bibl id="B16">
  <title><p>Semi-supervised clinical text classification with Laplacian SVMs:
  an application to cancer case management</p></title>
  <aug>
    <au><snm>Garla</snm><fnm>V</fnm></au>
    <au><snm>Taylor</snm><fnm>C</fnm></au>
    <au><snm>Brandt</snm><fnm>C</fnm></au>
  </aug>
  <source>Journal of biomedical informatics</source>
  <publisher>Elsevier</publisher>
  <pubdate>2013</pubdate>
  <volume>46</volume>
  <issue>5</issue>
  <fpage>869</fpage>
  <lpage>-875</lpage>
</bibl>

<bibl id="B17">
  <title><p>Active learning for clinical text classification: is it better than
  random sampling?</p></title>
  <aug>
    <au><snm>Figueroa</snm><fnm>RL</fnm></au>
    <au><snm>Zeng Treitler</snm><fnm>Q</fnm></au>
    <au><snm>Ngo</snm><fnm>LH</fnm></au>
    <au><snm>Goryachev</snm><fnm>S</fnm></au>
    <au><snm>Wiechmann</snm><fnm>EP</fnm></au>
  </aug>
  <source>Journal of the American Medical Informatics Association</source>
  <publisher>The Oxford University Press</publisher>
  <pubdate>2012</pubdate>
  <volume>19</volume>
  <issue>5</issue>
  <fpage>809</fpage>
  <lpage>-816</lpage>
</bibl>

<bibl id="B18">
  <title><p>Convolutional Neural Networks for Sentence
  Classification</p></title>
  <aug>
    <au><snm>Kim</snm><fnm>Y</fnm></au>
  </aug>
  <source>Proceedings of the 2014 Conference on Empirical Methods in Natural
  Language Processing (EMNLP)</source>
  <pubdate>2014</pubdate>
  <fpage>1746</fpage>
  <lpage>-1751</lpage>
</bibl>

<bibl id="B19">
  <title><p>A Convolutional Neural Network for Modelling Sentences</p></title>
  <aug>
    <au><snm>Kalchbrenner</snm><fnm>N</fnm></au>
    <au><snm>Grefenstette</snm><fnm>E</fnm></au>
    <au><snm>Blunsom</snm><fnm>P</fnm></au>
  </aug>
  <source>Proceedings of the 52nd Annual Meeting of the Association for
  Computational Linguistics (Volume 1: Long Papers)</source>
  <pubdate>2014</pubdate>
  <volume>1</volume>
  <fpage>655</fpage>
  <lpage>-665</lpage>
</bibl>

<bibl id="B20">
  <title><p>Improved Semantic Representations From Tree-Structured Long
  Short-Term Memory Networks</p></title>
  <aug>
    <au><snm>Tai</snm><fnm>KS</fnm></au>
    <au><snm>Socher</snm><fnm>R</fnm></au>
    <au><snm>Manning</snm><fnm>CD</fnm></au>
  </aug>
  <source>Proceedings of the 53rd Annual Meeting of the Association for
  Computational Linguistics and the 7th International Joint Conference on
  Natural Language Processing (Volume 1: Long Papers)</source>
  <pubdate>2015</pubdate>
  <volume>1</volume>
  <fpage>1556</fpage>
  <lpage>-1566</lpage>
</bibl>

<bibl id="B21">
  <title><p>Hierarchical attention networks for document
  classification</p></title>
  <aug>
    <au><snm>Yang</snm><fnm>Z</fnm></au>
    <au><snm>Yang</snm><fnm>D</fnm></au>
    <au><snm>Dyer</snm><fnm>C</fnm></au>
    <au><snm>He</snm><fnm>X</fnm></au>
    <au><snm>Smola</snm><fnm>A</fnm></au>
    <au><snm>Hovy</snm><fnm>E</fnm></au>
  </aug>
  <source>Proceedings of the 2016 Conference of the North American Chapter of
  the Association for Computational Linguistics: Human Language
  Technologies</source>
  <pubdate>2016</pubdate>
  <fpage>1480</fpage>
  <lpage>-1489</lpage>
</bibl>

<bibl id="B22">
  <title><p>Semi-supervised learning of the electronic health record for
  phenotype stratification</p></title>
  <aug>
    <au><snm>Beaulieu Jones</snm><fnm>BK</fnm></au>
    <au><snm>Greene</snm><fnm>CS</fnm></au>
    <au><cnm>others</cnm></au>
  </aug>
  <source>Journal of biomedical informatics</source>
  <publisher>Elsevier</publisher>
  <pubdate>2016</pubdate>
  <volume>64</volume>
  <fpage>168</fpage>
  <lpage>-178</lpage>
</bibl>

<bibl id="B23">
  <title><p>Comparing Rule-Based and Deep Learning Models for Patient
  Phenotyping</p></title>
  <aug>
    <au><snm>Gehrmann</snm><fnm>S</fnm></au>
    <au><snm>Dernoncourt</snm><fnm>F</fnm></au>
    <au><snm>Li</snm><fnm>Y</fnm></au>
    <au><snm>Carlson</snm><fnm>ET</fnm></au>
    <au><snm>Wu</snm><fnm>JT</fnm></au>
    <au><snm>Welt</snm><fnm>J</fnm></au>
    <au><snm>Foote Jr</snm><fnm>J</fnm></au>
    <au><snm>Moseley</snm><fnm>ET</fnm></au>
    <au><snm>Grant</snm><fnm>DW</fnm></au>
    <au><snm>Tyler</snm><fnm>PD</fnm></au>
    <au><cnm>others</cnm></au>
  </aug>
  <source>arXiv preprint arXiv:1703.08705</source>
  <pubdate>2017</pubdate>
</bibl>

<bibl id="B24">
  <title><p>2010 i2b2/VA challenge on concepts, assertions, and relations in
  clinical text</p></title>
  <aug>
    <au><snm>Uzuner</snm><fnm>{\"O}</fnm></au>
    <au><snm>South</snm><fnm>BR</fnm></au>
    <au><snm>Shen</snm><fnm>S</fnm></au>
    <au><snm>DuVall</snm><fnm>SL</fnm></au>
  </aug>
  <source>Journal of the American Medical Informatics Association</source>
  <publisher>BMJ Group BMA House, Tavistock Square, London, WC1H
  9JR</publisher>
  <pubdate>2011</pubdate>
  <volume>18</volume>
  <issue>5</issue>
  <fpage>552</fpage>
  <lpage>-556</lpage>
</bibl>

<bibl id="B25">
  <title><p>Recurrent neural networks for classifying relations in clinical
  notes</p></title>
  <aug>
    <au><snm>Luo</snm><fnm>Y</fnm></au>
  </aug>
  <source>Journal of biomedical informatics</source>
  <publisher>Elsevier</publisher>
  <pubdate>2017</pubdate>
  <volume>72</volume>
  <fpage>85</fpage>
  <lpage>-95</lpage>
</bibl>

<bibl id="B26">
  <title><p>Segment convolutional neural networks (Seg-CNNs) for classifying
  relations in clinical notes</p></title>
  <aug>
    <au><snm>Luo</snm><fnm>Y</fnm></au>
    <au><snm>Cheng</snm><fnm>Y</fnm></au>
    <au><snm>Uzuner</snm><fnm>{\"O}</fnm></au>
    <au><snm>Szolovits</snm><fnm>P</fnm></au>
    <au><snm>Starren</snm><fnm>J</fnm></au>
  </aug>
  <source>Journal of the American Medical Informatics Association</source>
  <publisher>Oxford University Press</publisher>
  <pubdate>2017</pubdate>
  <volume>25</volume>
  <issue>1</issue>
  <fpage>93</fpage>
  <lpage>-98</lpage>
</bibl>

<bibl id="B27">
  <title><p>Named entity recognition in Chinese clinical text using deep neural
  network</p></title>
  <aug>
    <au><snm>Wu</snm><fnm>Y</fnm></au>
    <au><snm>Jiang</snm><fnm>M</fnm></au>
    <au><snm>Lei</snm><fnm>J</fnm></au>
    <au><snm>Xu</snm><fnm>H</fnm></au>
  </aug>
  <source>Studies in health technology and informatics</source>
  <publisher>NIH Public Access</publisher>
  <pubdate>2015</pubdate>
  <volume>216</volume>
  <fpage>624</fpage>
</bibl>

<bibl id="B28">
  <title><p>Applying deep neural networks to unstructured text notes in
  electronic medical records for phenotyping youth depression</p></title>
  <aug>
    <au><snm>Geraci</snm><fnm>J</fnm></au>
    <au><snm>Wilansky</snm><fnm>P</fnm></au>
    <au><snm>Luca</snm><fnm>V</fnm></au>
    <au><snm>Roy</snm><fnm>A</fnm></au>
    <au><snm>Kennedy</snm><fnm>JL</fnm></au>
    <au><snm>Strauss</snm><fnm>J</fnm></au>
  </aug>
  <source>Evidence-based mental health</source>
  <publisher>Royal College of Psychiatrists</publisher>
  <pubdate>2017</pubdate>
  <volume>20</volume>
  <issue>3</issue>
  <fpage>83</fpage>
  <lpage>-87</lpage>
</bibl>

<bibl id="B29">
  <title><p>Structured prediction models for RNN based sequence labeling in
  clinical text</p></title>
  <aug>
    <au><snm>Jagannatha</snm><fnm>AN</fnm></au>
    <au><snm>Yu</snm><fnm>H</fnm></au>
  </aug>
  <source>Proceedings of the Conference on Empirical Methods in Natural
  Language Processing. Conference on Empirical Methods in Natural Language
  Processing</source>
  <pubdate>2016</pubdate>
  <volume>2016</volume>
  <fpage>856</fpage>
</bibl>

<bibl id="B30">
  <title><p>Bidirectional RNN for medical event detection in electronic health
  records</p></title>
  <aug>
    <au><snm>Jagannatha</snm><fnm>AN</fnm></au>
    <au><snm>Yu</snm><fnm>H</fnm></au>
  </aug>
  <source>Proceedings of the conference. Association for Computational
  Linguistics. North American Chapter. Meeting</source>
  <pubdate>2016</pubdate>
  <volume>2016</volume>
  <fpage>473</fpage>
</bibl>

<bibl id="B31">
  <title><p>Learning to diagnose with LSTM recurrent neural
  networks</p></title>
  <aug>
    <au><snm>Lipton</snm><fnm>ZC</fnm></au>
    <au><snm>Kale</snm><fnm>DC</fnm></au>
    <au><snm>Elkan</snm><fnm>C</fnm></au>
    <au><snm>Wetzel</snm><fnm>R</fnm></au>
  </aug>
  <source>arXiv preprint arXiv:1511.03677</source>
  <pubdate>2015</pubdate>
</bibl>

<bibl id="B32">
  <title><p>Deep computational phenotyping</p></title>
  <aug>
    <au><snm>Che</snm><fnm>Z</fnm></au>
    <au><snm>Kale</snm><fnm>D</fnm></au>
    <au><snm>Li</snm><fnm>W</fnm></au>
    <au><snm>Bahadori</snm><fnm>MT</fnm></au>
    <au><snm>Liu</snm><fnm>Y</fnm></au>
  </aug>
  <source>Proceedings of the 21th ACM SIGKDD International Conference on
  Knowledge Discovery and Data Mining</source>
  <pubdate>2015</pubdate>
  <fpage>507</fpage>
  <lpage>-516</lpage>
</bibl>

<bibl id="B33">
  <title><p>Distributed representations of words and phrases and their
  compositionality</p></title>
  <aug>
    <au><snm>Mikolov</snm><fnm>T</fnm></au>
    <au><snm>Sutskever</snm><fnm>I</fnm></au>
    <au><snm>Chen</snm><fnm>K</fnm></au>
    <au><snm>Corrado</snm><fnm>GS</fnm></au>
    <au><snm>Dean</snm><fnm>J</fnm></au>
  </aug>
  <source>NIPS</source>
  <pubdate>2013</pubdate>
  <fpage>3111</fpage>
  <lpage>-3119</lpage>
</bibl>

<bibl id="B34">
  <title><p>MIMIC-III, a freely accessible critical care database</p></title>
  <aug>
    <au><snm>Johnson</snm><fnm>AE</fnm></au>
    <au><snm>Pollard</snm><fnm>TJ</fnm></au>
    <au><snm>Shen</snm><fnm>L</fnm></au>
    <au><snm>Li wei</snm><fnm>HL</fnm></au>
    <au><snm>Feng</snm><fnm>M</fnm></au>
    <au><snm>Ghassemi</snm><fnm>M</fnm></au>
    <au><snm>Moody</snm><fnm>B</fnm></au>
    <au><snm>Szolovits</snm><fnm>P</fnm></au>
    <au><snm>Celi</snm><fnm>LA</fnm></au>
    <au><snm>Mark</snm><fnm>RG</fnm></au>
  </aug>
  <source>Scientific data</source>
  <publisher>Nature Publishing Group</publisher>
  <pubdate>2016</pubdate>
  <volume>3</volume>
  <fpage>160035</fpage>
</bibl>

<bibl id="B35">
  <title><p>An overview of MetaMap: historical perspective and recent
  advances</p></title>
  <aug>
    <au><snm>Aronson</snm><fnm>AR</fnm></au>
    <au><snm>Lang</snm><fnm>FM</fnm></au>
  </aug>
  <source>Journal of the American Medical Informatics Association</source>
  <publisher>BMJ Group BMA House, Tavistock Square, London, WC1H
  9JR</publisher>
  <pubdate>2010</pubdate>
  <volume>17</volume>
  <issue>3</issue>
  <fpage>229</fpage>
  <lpage>-236</lpage>
</bibl>

<bibl id="B36">
  <title><p>Medical semantic similarity with a neural language
  model</p></title>
  <aug>
    <au><snm>De Vine</snm><fnm>L</fnm></au>
    <au><snm>Zuccon</snm><fnm>G</fnm></au>
    <au><snm>Koopman</snm><fnm>B</fnm></au>
    <au><snm>Sitbon</snm><fnm>L</fnm></au>
    <au><snm>Bruza</snm><fnm>P</fnm></au>
  </aug>
  <source>Proceedings of the 23rd ACM international conference on conference on
  information and knowledge management</source>
  <pubdate>2014</pubdate>
  <fpage>1819</fpage>
  <lpage>-1822</lpage>
</bibl>

<bibl id="B37">
  <title><p>TensorFlow: A System for Large-Scale Machine Learning.</p></title>
  <aug>
    <au><snm>Abadi</snm><fnm>M</fnm></au>
    <au><snm>Barham</snm><fnm>P</fnm></au>
    <au><snm>Chen</snm><fnm>J</fnm></au>
    <au><snm>Chen</snm><fnm>Z</fnm></au>
    <au><snm>Davis</snm><fnm>A</fnm></au>
    <au><snm>Dean</snm><fnm>J</fnm></au>
    <au><snm>Devin</snm><fnm>M</fnm></au>
    <au><snm>Ghemawat</snm><fnm>S</fnm></au>
    <au><snm>Irving</snm><fnm>G</fnm></au>
    <au><snm>Isard</snm><fnm>M</fnm></au>
    <au><cnm>others</cnm></au>
  </aug>
</bibl>

<bibl id="B38">
  <title><p>A method for stochastic optimization</p></title>
  <aug>
    <au><snm>Kinga</snm><fnm>D</fnm></au>
    <au><snm>Adam</snm><fnm>JB</fnm></au>
  </aug>
  <source>International Conference on Learning Representations (ICLR)</source>
  <pubdate>2015</pubdate>
</bibl>

<bibl id="B39">
  <title><p>Medical subdomain classification of clinical notes using a machine
  learning-based natural language processing approach</p></title>
  <aug>
    <au><snm>Weng</snm><fnm>WH</fnm></au>
    <au><snm>Wagholikar</snm><fnm>KB</fnm></au>
    <au><snm>McCray</snm><fnm>AT</fnm></au>
    <au><snm>Szolovits</snm><fnm>P</fnm></au>
    <au><snm>Chueh</snm><fnm>HC</fnm></au>
  </aug>
  <source>BMC medical informatics and decision making</source>
  <publisher>BioMed Central</publisher>
  <pubdate>2017</pubdate>
  <volume>17</volume>
  <issue>1</issue>
  <fpage>155</fpage>
</bibl>

<bibl id="B40">
  <title><p>Contralateral Breast Cancer Event Detection Using Nature Language
  Processing</p></title>
  <aug>
    <au><snm>Zeng</snm><fnm>Z</fnm></au>
    <au><snm>Li</snm><fnm>X</fnm></au>
    <au><snm>Espino</snm><fnm>S</fnm></au>
    <au><snm>Roy</snm><fnm>A</fnm></au>
    <au><snm>Kitsch</snm><fnm>K</fnm></au>
    <au><snm>Clare</snm><fnm>S</fnm></au>
    <au><snm>Khan</snm><fnm>S</fnm></au>
    <au><snm>Luo</snm><fnm>Y</fnm></au>
  </aug>
  <source>AMIA Annual Symposium Proceedings</source>
  <pubdate>2017</pubdate>
  <volume>2017</volume>
  <fpage>1885</fpage>
</bibl>

</refgrp>
} 

    \end{backmatter}
    \end{document}